\documentclass[]{TEAI}
\usepackage{helvet}
\usepackage[utf8]{inputenc}
\usepackage{comment,wrapfig,hhline,rotating,adjustbox}
\usepackage{tabularx,array,makecell,textcomp,enumitem,threeparttable}
\usepackage{colortbl,float,xspace,url}
\newtheorem{proposition}{Proposition}

\hypersetup{
  pdftitle={Search-G1: Co-Evolving Representation-Based Rewards for Selective Grounded Search},
  pdfauthor={Ruoxi Cheng, Haoxuan Ma, Hongyi Zhang, Junming Zhang, Ranjie Duan, Qiaolin Xia, Hao Wang, Yu Lu, Haibo Shi, Xingjun Ma}
}

\title{Search-G1: Co-Evolving Representation-Based Rewards for Selective Grounded Search}
\author{
Ruoxi Cheng\textsuperscript{1,2},
Haoxuan Ma\textsuperscript{3},
Hongyi Zhang\textsuperscript{4},
Junming Zhang\textsuperscript{5},\\[0.8mm]
Ranjie Duan\textsuperscript{2},
Qiaolin Xia\textsuperscript{2},
Hao Wang\textsuperscript{2,$\dagger$},
Yu Lu\textsuperscript{2},
Haibo Shi\textsuperscript{2},
Xingjun Ma\textsuperscript{1,$\dagger$}
}
\affiliation{
\textsuperscript{1}Fudan University\quad
\textsuperscript{2}Tencent\quad
\textsuperscript{3}Nanjing University\\
\textsuperscript{4}Nanyang Technological University\quad
\textsuperscript{5}Shanghai Jiao Tong University
}
\contribution[\dagger]{Corresponding authors}
\abstract{
Search-augmented language agents should retrieve external information only when necessary and ground their answers in retrieved evidence. Yet terminal correctness assigns equal credit to direct answers from parametric knowledge, evidence-grounded answers, and answers recalled after redundant search. Process supervision and LLM judges provide richer feedback but can require high-cost annotation or inference; low-cost, non-contrastive policy confidence signals do not directly identify whether retrieved evidence changes an answer.
We propose \textbf{Search-G1}, a reward framework that couples retrieval necessity with evidence reliance through two intervention-calibrated hidden-state readouts for \textbf{selective grounded search}. A prompt readout predicts closed-book sufficiency, whose complement proxies policy-relative \textbf{retrieval necessity}; an answer-commit readout predicts \textbf{evidence reliance} as sensitivity to evidence deletion conditional on the realized reasoning prefix. A correctness-first reward gates evidence-reliance credit by retrieval necessity, rewards sufficient direct answers, and penalizes repeated search. After calibration, scoring needs neither process annotations nor LLM-judge inference.
As policy-relative targets, trajectory distributions, and representations may drift, Search-G1 periodically regenerates targets and refits its readouts. Across four QA benchmarks and two scales, Search-G1 achieves competitive task accuracy, with stronger counterfactual evidence dependence and fewer searches in the main comparisons, complemented by independent human audits of answer support.
}
\correspondence{\email{vichwang@tencent.com}; \email{xingjunma@fudan.edu.cn}}
\checkdata[Code]{\url{https://github.com/Rosy0912/Search-G1}}

\begin{document}
\raggedbottom
\maketitle
\renewcommand{\thefootnote}{\fnsymbol{footnote}}
\footnotetext[2]{This work was conducted during Ruoxi Cheng's participation in the Tencent (Yuanbao AI Search) Rhino-Bird Research Elite Program (\textit{Industrial Supervisor: Hao Wang; Academic Supervisor: Xingjun Ma}).}
\renewcommand{\thefootnote}{\arabic{footnote}}

\section{Introduction}

Search agents augment large language models (LLMs) with an external retrieval loop. They issue queries, inspect documents, reason over evidence, and then commit an answer~\citep{shang2025agentsquare,zhan2026safesearch}. Recent search-agent systems show that reinforcement learning can induce this behavior~\citep{jin2025search,song2025r1,gao2025beyond}. Yet the interface only determines what an agent can do. The reward determines which behavior policy optimization reinforces. Terminal correctness cannot distinguish grounded retrieval from redundant search followed by parametric recall, so trajectories that differ sharply in search behavior are reinforced identically. A capable agent should instead perform \emph{selective grounded search}: retrieve when needed, and ground the answer in what was retrieved. This requires two distinct judgments. \emph{Retrieval necessity}~\citep{feng2026trust} asks whether the current policy can answer without retrieval. \emph{Evidence reliance} asks whether a searched answer actually depends on the retrieved context~\citep{yin2026deliberative}. Rewarding reliance alone can favor redundant retrieval; rewarding necessity alone does not distinguish evidence use from recall after search.

\begin{figure}[t]
\centering
\includegraphics[width=0.72\linewidth]{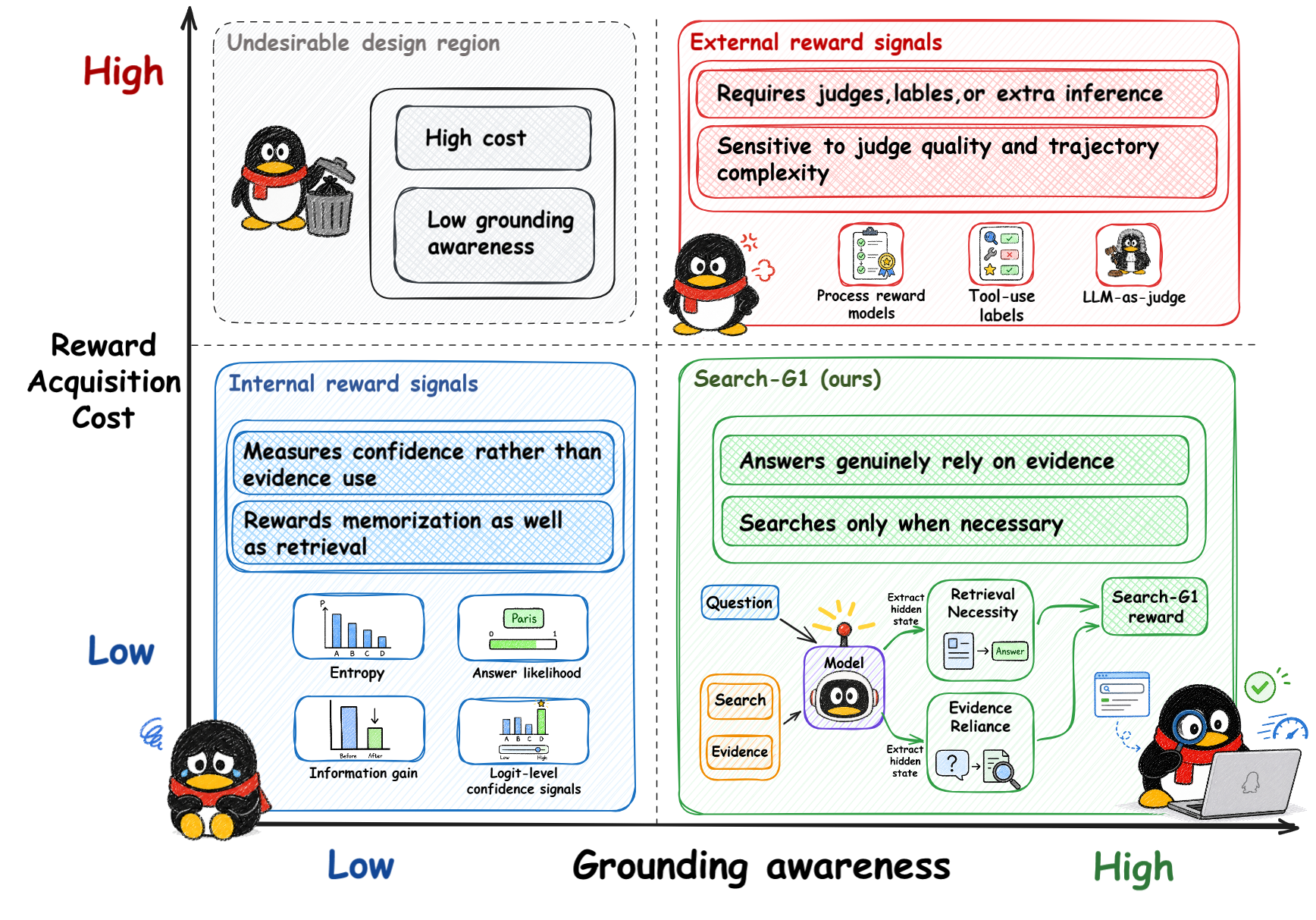}
\caption{Reward signals for agentic search RL.
External rewards range from sparse outcomes to process supervision, tool feedback, and judge signals.
Internal rewards provide graded signals but often capture confidence rather than evidence use.
These families differ in the trajectory-level credit they assign to correct search behavior: trajectories that are equally correct may warrant different credit depending on whether retrieval was necessary and whether the answer used the retrieved evidence.
Search-G1 calibrates two policy readouts through interventions: a prompt readout estimates whether the policy can answer without retrieval, and an answer readout estimates reliance on retrieved evidence.}
\label{fig:intro}
\end{figure}

Retrieval need depends on both question structure and policy knowledge~\citep{zhao2026pre}. Our initial Qwen2.5-3B-Instruct policy~\citep{ahmed2025qwen} answers $14\%$ of held-out NQ~\citep{kwiatkowski2019natural} questions correctly without retrieval, suggesting scope to avoid redundant search. HotpotQA~\citep{yang2018hotpotqa} has $79.9\%$ bridge questions, and MuSiQue~\citep{trivedi2022musique} contains $2$--$4$-hop questions (\Cref{tab:boundary_composition}). These structural annotations characterize task demands, not the evidence a particular policy needs. We therefore make both judgments policy-relative: necessity is derived from closed-book sufficiency, while reliance measures answer-stage sensitivity to evidence deletion conditional on the realized reasoning prefix (\Cref{sec:method_calibration}).

Existing reward designs fall into two broad families, as shown in \Cref{fig:intro}. \textbf{External rewards} include sparse answer correctness~\citep{song2025r1,jin2025search}, process reward models~\citep{wang2025spa,xi2026agentprm}, tool-grounding feedback~\citep{ma2025pou}, and LLM-critic or learned-reward feedback~\citep{zhang2026criticsearch,hu2025openreward}. Richer external signals inspect trajectories more directly, but require annotations or additional model calls and may be brittle on multi-step traces~\citep{liu2026agenthallu}. \textbf{Internal rewards} instead derive graded signals from the policy itself~\citep{du2019liir}, including entropy~\citep{dong2026agentic}, answer likelihood or information gain~\citep{wang2026information}, and logit-level confidence. Non-contrastive confidence signals do not directly identify evidence use: a model may remain confident because it already knows an answer, even after searching~\citep{mallen2023not,wang2023survey,tang2026saas}. Contrastive context-masking rewards assess evidence sensitivity more directly~\citep{tan2026ctrl,guo2026contextual}. We target a reward that jointly measures retrieval necessity and evidence reliance, with the underlying interventions amortized into readouts that can be refreshed as the policy changes.

To address this gap, we introduce \textbf{Search-G1}, a policy-relative reward framework for selective grounded search that combines two intervention-calibrated readouts of the policy's internal representations. A prompt-state readout predicts closed-book sufficiency, whose complement gives retrieval necessity; an answer-commit readout estimates evidence reliance from sensitivity to evidence deletion. Both readouts amortize these interventions during rollout scoring, requiring neither process annotations nor external judge calls at reward time. Their scores form a correctness-first trajectory reward: searched answers receive necessity-conditioned reliance credit, direct answers receive sufficiency credit, and repeated search incurs a bounded cost. This design can restore variation among equally correct trajectories that differ in reliance, branch, or cost.
Because reinforcement learning changes policy-relative targets, trajectories, and checkpoint representations, fixed readouts can become stale. Search-G1 therefore periodically regenerates targets and refits both readouts: interventions calibrate, the reward assigns credit, masked GRPO updates the policy, and the next round recalibrates on the latest checkpoint. Across four QA benchmarks and two scales, Search-G1 achieves competitive task accuracy, with stronger counterfactual evidence dependence and fewer searches than other search-enabled baselines in the main grounding evaluations. Independent human-support and fixed-question audits complement these results.

Our main contributions are as follows:
\begin{itemize}

    \item We formulate selective grounded search as a \textbf{policy-relative reward problem}: distinguishing equally correct trajectories requires judging whether retrieval was necessary for the current policy and whether the answer relied on the retrieved evidence, neither of which is available from terminal correctness.

    \item We propose \textbf{Search-G1}, which distills question-level closed-book sufficiency and trajectory-level evidence-deletion sensitivity into internal readouts. A \textbf{necessity-gated reward} conditions evidence-reliance credit on closed-book insufficiency while rewarding sufficient direct answers; periodic target regeneration and refitting adapt both readouts to the current policy.

    \item Experiments show competitive task accuracy, with stronger counterfactual evidence dependence and fewer searches in the main comparisons. Human-support and fixed-question analyses complement these diagnostic results.
\end{itemize}

\section{Related Work}
\subsection{Search-Agent Rewards and Credit Assignment}
Self-RAG uses reflection tokens for retrieval and support judgments~\citep{asai2024self}, while Adaptive-RAG routes questions by complexity~\citep{jeong2024adaptive}. Search-agent RL instead assigns credit to sampled trajectories. Outcome-driven agents use terminal correctness in Search-R1~\citep{jin2025search} and longer-horizon training in Beyond Ten Turns~\citep{gao2025beyond}. Denser credit comes from AgentPRM step scores~\citep{xi2026agentprm}, GiGPO anchor-state advantages~\citep{feng2026group}, or the CriticSearch retrospective critic~\citep{zhang2026criticsearch}; PoU checks citation validity, perturbation sensitivity, and answer--evidence alignment~\citep{ma2025pou}. Internal shaping uses post-tool entropy~\citep{dong2026agentic}, gold-answer information gain~\citep{wang2026information}, or sibling-context distillation~\citep{liang2026search}. Boundary methods use query confidence~\citep{wu2025search}, certainty plus retrieval quality~\citep{feng2026trust}, disabled/enabled rollout contrasts~\citep{tang2026saas}, or AKBE's dynamic tool/no-tool probes~\citep{chen2026efficient}. CTRL-RAG contrasts response likelihoods with and without supporting documents~\citep{tan2026ctrl}, while concurrent CIPO masks attention to retrieved observations to score subsequent actions~\citep{guo2026contextual}. Concurrent work co-evolves textual guidance~\citep{wang2026co} or critic supervision~\citep{zhang2026self}. Search-G1 makes current-policy retrieval necessity an explicit gate on evidence-reliance credit, using periodically refreshed hidden-state readouts to amortize retrieval-disabled and evidence-deletion interventions.

\subsection{Internal Signals of Evidence Dependence}
Internal-state studies examine factual knowledge and its use in context, including relational facts~\citep{petroni2019language}, key--value memories~\citep{geva2021transformer}, factual knowledge representations~\citep{hernandez2023inspecting}, and knowledge--context reconciliation~\citep{zhao2026understanding}. Related signals support correctness self-assessment~\citep{kadavath2022language}, truthfulness prediction~\citep{azaria2023internal}, entity-knowledge estimation~\citep{ferrando2025know}, and tool-selection error detection~\citep{healy2026internal}. Applications include RAG attribution~\citep{qi2024model}, hallucination detection~\citep{sun2025redeep}, and factuality decoding~\citep{chuang2024dola}. Using these signals as training rewards also requires accounting for the policy that produces them. Most of the above methods target fixed-model inference; representation-based~\citep{cheng2026adversarial} exploration extends internal signals to post-training~\citep{tuyls2026representation}, while PAIR combines a frozen belief-consistency probe with attention correction~\citep{kim2026pair}. Search-G1 calibrates retrieval-specific targets for closed-book sufficiency and evidence-deletion sensitivity and repeatedly refits both policy-conditioned readouts.

\section{Methodology}
\label{sec:methodology}
\begin{figure}[t]
    \centering
\includegraphics[width=0.98\linewidth]{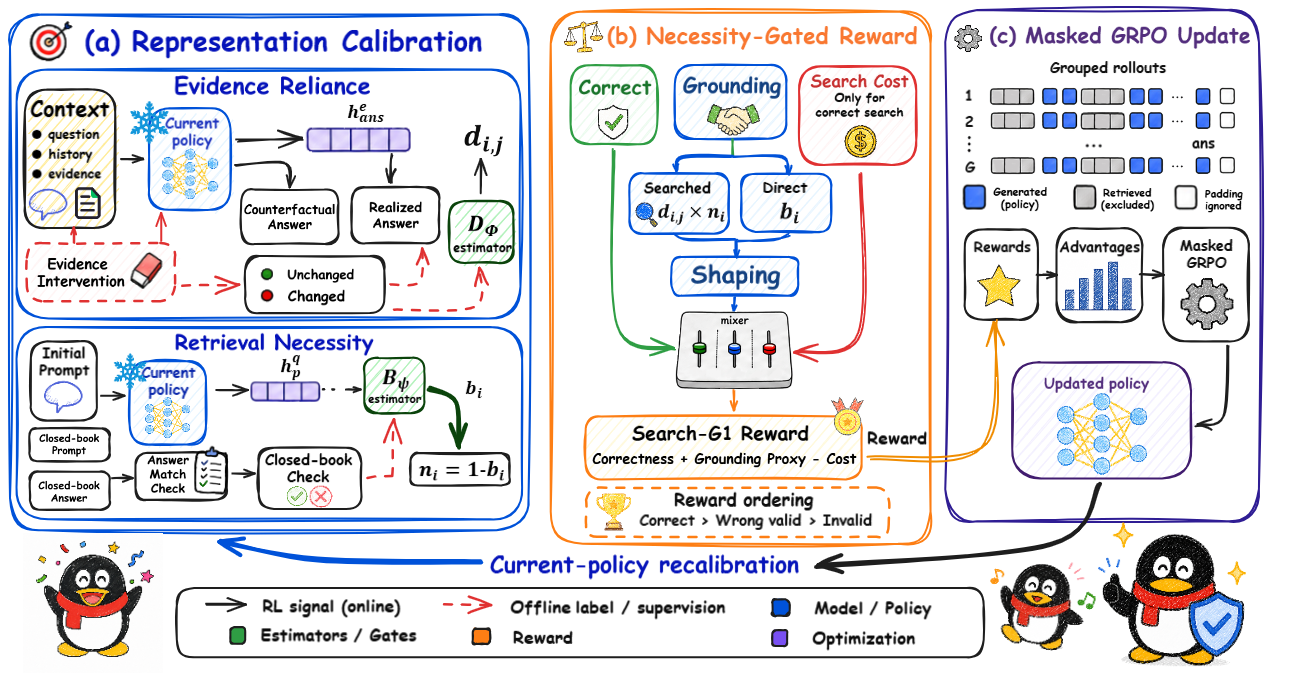}
    \caption{Pipeline of \textbf{Search-G1}. We first calibrate two hidden-state readouts from closed-book and evidence-deletion interventions on a frozen policy snapshot. We then combine their scores into necessity-gated trajectory rewards and update the policy with masked GRPO. Finally, we periodically refresh the snapshot, regenerate intervention targets, and refit both readouts as the policy evolves.}
    \label{fig:Search-G1_pipeline}
\end{figure}

In this section, we present Search-G1, a reward framework for selective grounded search as shown in \Cref{fig:Search-G1_pipeline}. We first define the intervention targets and calibrate two representation readouts. Then, we combine their scores into a necessity-gated trajectory reward. Finally, we describe the masked GRPO update, its normalization effects, and periodic readout recalibration.

\subsection{Problem Formulation}
\label{sec:method_problem}
Given a question $q_i$, gold answer $y_i^\star$, and search-enabled prompt $P_i$, the rollout policy samples a group of $G\ge2$ trajectories,
\[
\tau_{i,j}\overset{\mathrm{iid}}{\sim}
\pi_{\mathrm{old}}(\cdot\mid P_i),
\qquad j=1,\ldots,G.
\]
The rollout policy $\pi_{\mathrm{old}}$ remains fixed while $\pi_\theta$ is updated on the sampled group. Each trajectory contains reasoning, search actions, retrieved observations, and a final answer $\hat a_{i,j}$. Let $v_{i,j}\in\{0,1\}$ indicate validity under the output parser, and let $e_{i,j}=V(\hat a_{i,j},y_i^\star)\in\{0,1\}$ indicate answer correctness, with $e_{i,j}=0$ for invalid outputs. The statistic $N_{i,j}\in\mathbb Z_{\ge0}$ counts parsed search actions: $N_{i,j}=0$ denotes a direct trajectory and $N_{i,j}>0$ a searched one.

The goal is to convert policy-relative retrieval necessity and evidence sensitivity into correctness-first trajectory credit: favor evidence-sensitive search when needed, direct answering when closed-book knowledge suffices, and discourage repeated search. Because GRPO normalizes within each rollout group, the reward is also designed to differentiate trajectories that a correctness-only signal would tie.

\subsection{Intervention-Calibrated Representation Readouts}
\label{sec:method_calibration}
The two readouts operate at different levels: closed-book sufficiency is shared across a question's rollouts, whereas evidence reliance depends on each searched trajectory. At calibration round $m$, we freeze the latest checkpoint $\bar\theta_m$. It generates counterfactual completions and encodes the prompt-end state $h_{p,i}$ and final-answer state $h_{\mathrm{ans},i,j}$ from original, unperturbed trajectories. Between refits, current trajectories are teacher-forced through the frozen evaluator without gradients; the snapshot and both heads stay fixed, and one evaluator version scores each rollout group.

\paragraph{Evidence Reliance.}
For a valid searched trajectory, let $C_{i,j}$ be its context immediately before the answer. The intervention $T(C_{i,j})$ removes retrieved-observation blocks while holding the question, search history, and realized reasoning fixed, measuring evidence sensitivity conditional on this prefix. We define
\begin{equation}
\label{eq:evidence_reliance_def}
\begin{aligned}
s_{i,j}^{\mathrm{chg}}
&=
\mathbf 1\!\left[
\nu(\hat a_{i,j})
\neq
\nu\!\left(A_{\bar\theta_m}^{\mathrm{det}}(T(C_{i,j}))\right)
\right],
\\
d_{i,j}
&=D_{\phi^{(m)}}(h_{\mathrm{ans},i,j})\in[0,1],
\end{aligned}
\end{equation}
where $A_{\bar\theta_m}^{\mathrm{det}}(C)$ is the parsed answer from a greedy continuation of prefix $C$ under the snapshot, and $\nu$ applies the task evaluator's deterministic string normalization (lowercasing, article and punctuation removal, and whitespace collapsing). The target compares a sampled answer with a greedy counterfactual; \Cref{tab:label_audit} audits decoding mismatch. The readout estimates deletion sensitivity and contributes reward only in the correct-searched branch of~\Cref{eq:searchg1_reward}.

\paragraph{Retrieval Necessity.}
From a fixed retrieval-disabled prompt $P_i^{\mathrm{cb}}$, we obtain a closed-book target and predict it from the actual search-enabled prompt state:
\begin{equation}
\label{eq:necessity_target}
\begin{aligned}
z_i
&=V_{\mathrm{cb}}\!\left(
A_{\bar\theta_m}^{\mathrm{det}}(P_i^{\mathrm{cb}}),
y_i^\star
\right)\in\{0,1\},
\\
b_i
&=B_{\psi^{(m)}}(h_{p,i})\in[0,1],
\qquad n_i=1-b_i.
\end{aligned}
\end{equation}
Here $b_i$ estimates sufficiency under the frozen policy, closed-book prompt, deterministic decoder, and verifier; $n_i$ is therefore policy- and protocol-relative. The target is defined by the retrieval-disabled prompt, while the readout acts on the state that $P_i$ induces before any search. Thus $b_i$ predicts current-policy closed-book sufficiency from the decision state and provides a graded reward-time score without an auxiliary closed-book completion. $V_{\mathrm{cb}}$ uses normalized inclusion for free-form completions, which is more permissive than the exact-match criterion used for outcomes; disagreements shift targets toward sufficiency and may bias credit toward direct behavior. Exact conventions are in~\Cref{app:audits}.

Let $\mathcal D_D^{(m)}$ contain valid searched trajectories and $\mathcal D_B^{(m)}$ contain question-level closed-book examples. The two readouts are fitted with
\begin{equation}
\label{eq:refit_loss}
\mathcal L_{\mathrm{cal}}
=
\mathbb E_{\mathcal D_D^{(m)}}
\!\left[\mathrm{BCE}(d_{i,j},s_{i,j}^{\mathrm{chg}})\right]
+
\mathbb E_{\mathcal D_B^{(m)}}
\!\left[\mathrm{BCE}(b_i,z_i)\right].
\end{equation}
Hard targets are generated on the disjoint calibration pool. The fitted readouts map rollout states to continuous scores and supply both signals in one reward-time forward pass, amortizing a closed-book decode per distinct scored question and an evidence-deleted decode per valid searched trajectory. At each refit, each head selects its layer by validation AUC; fitting, refresh overhead, and label construction are detailed in~\Cref{app:implementation_details,app:analysis,app:audits}.

\subsection{Necessity-Gated Reward}
\label{sec:method_reward}
For a searched trajectory, Search-G1 combines evidence reliance and retrieval necessity into a necessity-gated reliance score
\begin{equation}
\label{eq:grounding_score}
g_{i,j}=d_{i,j}n_i.
\end{equation}
This multiplicative gate is large only when the answer is evidence-sensitive and closed-book sufficiency is low. The product $g_{i,j}$ serves as a graded shaping score for correct searched trajectories. Correct direct trajectories instead use $b_i$.

Within the correct-searched reward branch, the first search is unpenalized, while any trajectory with two or more search actions activates a unit cost:
\begin{equation}
\label{eq:search_cost}
c_{i,j}=\mathbf 1[N_{i,j}>1].
\end{equation}
Direct and single-search trajectories have zero cost; repeated search has unit cost regardless of further calls. Under the four-round cap, this bounded term distinguishes single from repeated retrieval, preserving the correct-searched lower bound of $1-\eta$ and the correctness-first reward tiers.

Let $r_{\mathrm{inv}}$ and $r_{\mathrm{wrong}}$ denote the rewards for invalid and wrong outputs. The Search-G1 reward is
\begin{equation}
\label{eq:searchg1_reward}
r_{i,j}
=
\begin{cases}
r_{\mathrm{inv}},
& v_{i,j}=0,
\\[2pt]
r_{\mathrm{wrong}},
& v_{i,j}=1,\ e_{i,j}=0,
\\[2pt]
1+\lambda_g g_{i,j}-\eta c_{i,j},
& v_{i,j}=1,\ e_{i,j}=1,\ N_{i,j}>0,
\\[2pt]
1+\alpha_{\mathrm{cb}}b_i,
& v_{i,j}=1,\ e_{i,j}=1,\ N_{i,j}=0.
\end{cases}
\end{equation}

For a correct searched trajectory and a correct direct trajectory from the same question, the searched trajectory ranks higher before---and therefore after---group normalization exactly when
\begin{equation}
\label{eq:direct_search_order}
\lambda_g d_{i,j}(1-b_i)-\eta c_{i,j}
>
\alpha_{\mathrm{cb}}b_i.
\end{equation}
Higher closed-book sufficiency thus shifts credit toward direct answering, while reliance and necessity must jointly offset search cost; refitting the sufficiency readout against the latest checkpoint keeps this boundary tied to current capability. The tiers themselves hold whenever $r_{\mathrm{inv}}<r_{\mathrm{wrong}}<1-\eta$, $\eta\in[0,1)$, and $\lambda_g,\alpha_{\mathrm{cb}}\ge0$: every correct valid trajectory outranks every wrong one, which outranks every invalid output. Nonnegativity is all this requires; our coefficient values are listed in~\Cref{app:implementation_details}.

\subsection{Group-Relative Credit under Masked GRPO}
\label{sec:method_optimization}

Under masked GRPO, each trajectory receives the group-relative advantage~\citep{shao2024deepseekmath},
\begin{equation}
\label{eq:group_advantage}
\hat A_{i,j}
=
\frac{r_{i,j}-\mu_i}{\sigma_i+\delta},
\end{equation}
where $\mu_i$ and $\sigma_i$ are the group mean and sample standard deviation and $\delta>0$ stabilizes normalization.

We use the standard clipped GRPO update with Search-G1 rewards. Let $M_{i,j,t}=1$ for reasoning, search-action, and final-answer tokens and $0$ for prompts, retrieved observations, and padding. We write $\langle\cdot\rangle_{\mathcal M,\mathcal B}$ for the masked token average over batch $\mathcal B$ and maximize
\begin{equation}
\label{eq:masked_grpo}
\begin{aligned}
\ell^{\mathrm{clip}}_{i,j,t}
={}&
\min\!\left\{
\rho_{i,j,t}\hat A_{i,j},
\operatorname{clip}(\rho_{i,j,t},1-\epsilon,1+\epsilon)
\hat A_{i,j}
\right\},
\\
\mathcal J_{\mathrm{G1}}(\theta)
={}&
\left\langle
\ell^{\mathrm{clip}}_{i,j,t}
+\alpha_{\mathrm H}\mathcal H(\pi_\theta(\cdot\mid\tau_{i,j,<t}))
-\beta_{\mathrm{KL}}\widehat{\mathrm{KL}}^{\mathrm{LV}}_{i,j,t}
\right\rangle_{\mathcal M,\mathcal B},
\end{aligned}
\end{equation}
Here $\rho_{i,j,t}$ is the current-to-rollout-policy token likelihood ratio, $\epsilon$ is the clipping width, and $\alpha_{\mathrm H},\beta_{\mathrm{KL}}$ weight entropy and reference-policy KL regularization. The reward enters through $\hat A_{i,j}$, so its within-group distinctions determine the credit assigned to policy-generated tokens. The ratio, KL estimator, and distributed averaging rule are specified in~\Cref{app:methodology_supp}.

\paragraph{Effect of GRPO Normalization.}
Centering removes group-wide additive terms, so the group-constant $n_i$ and $b_i$ act differently from the trajectory-varying $d_{i,j}$: a multiplicative gate can still reweight searched credit against the other branches. To characterize this, let $\gamma\in[0,1]$ replace $n_i$ in the correct-searched reliance term only, so that $\gamma=n_i$ recovers Search-G1 and $\gamma=1$ removes the gate; let $\mathbf u_i$ and $\mathbf v_i$ be the group-centered versions of the $\gamma=0$ reward vector and of $x_{i,j}=d_{i,j}\mathbf 1[v_{i,j}=e_{i,j}=1,\,N_{i,j}>0]$, and set $\mathbf z_i(\gamma)=\mathbf u_i+\lambda_g\gamma\mathbf v_i$.

\begin{proposition}[Normalization property of the necessity gate]
\label{prop:gate_normalization}
Assume $\lambda_g>0$. For any $\gamma\in[0,1]$ with $\mathbf z_i(\gamma)\neq\mathbf 0$ and any $\delta\ge0$, group normalization yields
\begin{equation}
\label{eq:gated_advantage}
\mathbf A_i^\delta(\gamma)
=
\frac{\mathbf z_i(\gamma)}
{\|\mathbf z_i(\gamma)\|_2/\sqrt{G-1}+\delta},
\end{equation}
which preserves the direction of $\mathbf z_i(\gamma)$; when $\delta=0$, normalization also removes its scale. Consequently:
\emph{(i)} in an all-valid, all-correct, all-searched, equal-cost group with $\mathbf v_i\neq\mathbf 0$, $\mathbf u_i=\mathbf 0$, so every $\gamma>0$ gives the same rollout ranking, advantage signs, and credit direction, determined by $d_{i,j}$;
\emph{(ii)} for distinct $\gamma_1,\gamma_2$ with $\mathbf z_i(\gamma_1),\mathbf z_i(\gamma_2)\neq\mathbf 0$, the normalized credit directions differ whenever $\mathbf u_i$ and $\mathbf v_i$ are non-collinear.
\end{proposition}

This division of labor is what~\Cref{eq:searchg1_reward} implements: $d_{i,j}$ and $c_{i,j}$ separate trajectories inside the correct-searched branch, while $n_i$ and its complement $b_i$ act across branches, making the direct--search comparison monotone in closed-book sufficiency whenever both behaviors occur~(\Cref{eq:direct_search_order}). If $\mathbf z_i(\gamma)=\mathbf 0$, the centered reward is constant and yields no within-group advantage; \Cref{app:gate_diag} gives the proof, exact sign-change criterion, and boundary cases.

\paragraph{Current-Policy Recalibration.}
At each round, fixed calibration IDs are rerun under $\bar\theta_m$ to regenerate trajectories, targets, and states before both heads are refit and installed atomically. Here $\pi_{\mathrm{old}}$ is batch-local, whereas $\bar\theta_m$ persists across updates. Each refresh updates the intervention targets, their hidden-state representations, and the fitted heads, adapting both measurements to the latest policy. Schedules and overhead are given in~\Cref{app:implementation_details}.

\section{Evaluation}
\label{sec:evaluation}

We evaluate the overall accuracy--grounding--search trade-off, then examine whether the readouts capture their intended targets and provide useful credit as the policy evolves. Component ablations test the roles of the reward terms and evaluator refresh.

\subsection{Experimental Setup}
\label{sec:setup}

\paragraph{Models and retrieval environment.}
We evaluate Qwen2.5-3B/7B-Instruct~\citep{ahmed2025qwen} in a fixed Wikipedia-2018 retrieval environment with \texttt{e5-base-v2}, FAISS, three passages per retrieval, and at most four rounds. Training/calibration, validation/architecture-selection, and final-evaluation IDs are disjoint; validation selects reward and refresh settings, and checkpoints are fixed before held-out greedy evaluation without test-based selection, as detailed in~\Cref{app:implementation_details}.

\paragraph{Benchmarks and metrics.}
We report exact match (\textbf{EM}$\uparrow$) on NQ~\citep{kwiatkowski2019natural}, HotpotQA~\citep{yang2018hotpotqa}, 2WikiMultiHopQA~\citep{ho2020constructing}, and MuSiQue~\citep{trivedi2022musique}. On NQ and 2Wiki, grounding evaluation replaces an answer-bearing fact for originally correct searched trajectories. Searched-only \textbf{D-Inv}$\downarrow$ is the fraction preserving the original answer; \textbf{TC}$\uparrow$ combines correct-direct retention with graded searched-answer response; and \textbf{Search/Q}$\downarrow$ counts generated search markers per question. These diagnostics are not used as rollout rewards or readout-fitting targets; their validation-split values are used only for configuration selection. These diagnostics measure counterfactual evidence dependence. Because TC and D-Inv condition on policy outcomes, we complement them with searched-only accounting on fixed IDs and an unconditional common-question decomposition (\Cref{tab:efficiency_controlled,tab:decomposition}), alongside surface controls and independent human support. \Cref{tab:main,tab:grounding_2wiki} report NQ/2Wiki results; \Cref{app:analysis} defines the metrics.

\definecolor{SGGainGreen}{HTML}{2E7D32}
\definecolor{SGLossRed}{HTML}{C62828}

\newcommand{\sgMean}[2]{%
  \ensuremath{#1}}
\newcommand{\sgBest}[2]{%
  \ensuremath{\mathbf{#1}}}

\newcommand{\sgDelta}[3]{%
  {\textcolor{#1}{\scriptsize\ensuremath{#2\!#3}}}}
\newcommand{\sgGoodUp}[1]{\sgDelta{SGGainGreen}{\uparrow}{#1}}
\newcommand{\sgGoodDown}[1]{\sgDelta{SGGainGreen}{\downarrow}{#1}}
\newcommand{\sgBadUp}[1]{\sgDelta{SGLossRed}{\uparrow}{#1}}
\newcommand{\sgBadDown}[1]{\sgDelta{SGLossRed}{\downarrow}{#1}}

\newcommand{\sgResult}[3]{\sgMean{#1}{#2}\hspace{0.15em}#3}
\newcommand{\sgBestResult}[3]{\sgBest{#1}{#2}\hspace{0.15em}#3}
\newcommand{\sgReference}[2]{%
  \sgMean{#1}{#2}\hspace{0.15em}{\scriptsize--}}

\begin{table}[!tbp]
\centering
\captionsetup{justification=raggedright,singlelinecheck=false}
\caption{Comparison of Search-G1 and baseline methods across task accuracy, grounding, and search cost at two model scales. EM columns are dataset-specific; TC, D-Inv, and Search/Q are measured on NQ. Base/Prompted Search are deterministic. Arrows show deltas from Prompted Search; TC/D-Inv are n/a for Base.}
\label{tab:main}

\setlength{\tabcolsep}{3.2pt}
\renewcommand{\arraystretch}{0.82}

\begin{adjustbox}{max width=\linewidth}
\begin{tabular}{ll|cccc|cc|c}
\toprule
\multirow{2}{*}{Method}
& \multirow{2}{*}{Signal}
& \multicolumn{4}{c|}{\textbf{Task Utility}}
& \multicolumn{2}{c|}{\textbf{Grounding (NQ)}}
& \multicolumn{1}{c}{\textbf{Search Cost (NQ)}} \\
\cmidrule(lr){3-6}
\cmidrule(lr){7-8}
\cmidrule(lr){9-9}
&
& NQ ($\uparrow$)
& HotpotQA ($\uparrow$)
& 2WikiMultiHopQA  ($\uparrow$)
& MuSiQue ($\uparrow$)
& TC ($\uparrow$)
& D-Inv ($\downarrow$)
& Search/Q ($\downarrow$) \\
\midrule

\multicolumn{9}{c}{\textbf{Qwen2.5-3B-Instruct}} \\
\midrule

Base
& None
& 0.142
& 0.153
& 0.212
& 0.029
& n/a
& n/a
& 0.000 \\
\addlinespace[1pt]

\rowcolor{gray!5}
Prompted Search
& Retrieval
& 0.250
& 0.199
& 0.238
& 0.057
& 0.930
& 0.069
& 2.215 \\
\cmidrule(lr){1-9}

Search-R1~\citep{jin2025search}
& Outcome+Direct
& \sgResult{0.352}{0.019}{\sgGoodUp{0.102}}
& \sgResult{0.248}{0.015}{\sgGoodUp{0.049}}
& \sgResult{0.267}{0.014}{\sgGoodUp{0.029}}
& \sgResult{0.087}{0.012}{\sgGoodUp{0.030}}
& \sgResult{0.917}{0.018}{\sgBadDown{0.013}}
& \sgResult{0.049}{0.011}{\sgGoodDown{0.020}}
& \sgResult{3.281}{0.21}{\sgBadUp{1.066}} \\

\rowcolor{gray!5}
GiGPO~\citep{feng2026group}
& State
& \sgResult{0.422}{0.021}{\sgGoodUp{0.172}}
& \sgResult{0.296}{0.017}{\sgGoodUp{0.097}}
& \sgResult{0.330}{0.016}{\sgGoodUp{0.092}}
& \sgResult{0.126}{0.014}{\sgGoodUp{0.069}}
& \sgResult{0.935}{0.015}{\sgGoodUp{0.005}}
& \sgResult{0.047}{0.010}{\sgGoodDown{0.022}}
& \sgResult{3.131}{0.18}{\sgBadUp{0.916}} \\

AgentPRM~\citep{xi2026agentprm}
& Process
& \sgResult{0.380}{0.020}{\sgGoodUp{0.130}}
& \sgResult{0.293}{0.018}{\sgGoodUp{0.094}}
& \sgResult{0.328}{0.017}{\sgGoodUp{0.090}}
& \sgResult{0.112}{0.013}{\sgGoodUp{0.055}}
& \sgResult{0.958}{0.012}{\sgGoodUp{0.028}}
& \sgResult{0.042}{0.009}{\sgGoodDown{0.027}}
& \sgResult{3.340}{0.23}{\sgBadUp{1.125}} \\

\rowcolor{gray!5}
GPT-5 Judge~\citep{singh2025openai}
& External Judge
& \sgResult{0.402}{0.022}{\sgGoodUp{0.152}}
& \sgResult{0.280}{0.018}{\sgGoodUp{0.081}}
& \sgResult{0.324}{0.017}{\sgGoodUp{0.086}}
& \sgResult{0.101}{0.014}{\sgGoodUp{0.044}}
& \sgResult{0.946}{0.016}{\sgGoodUp{0.016}}
& \sgResult{0.038}{0.008}{\sgGoodDown{0.031}}
& \sgResult{3.238}{0.22}{\sgBadUp{1.023}} \\

ARPO~\citep{dong2026agentic}
& Entropy
& \sgResult{0.389}{0.018}{\sgGoodUp{0.139}}
& \sgResult{0.273}{0.016}{\sgGoodUp{0.074}}
& \sgResult{0.304}{0.015}{\sgGoodUp{0.066}}
& \sgResult{0.109}{0.013}{\sgGoodUp{0.052}}
& \sgResult{0.924}{0.017}{\sgBadDown{0.006}}
& \sgResult{0.055}{0.012}{\sgGoodDown{0.014}}
& \sgResult{2.679}{0.16}{\sgBadUp{0.464}} \\

\rowcolor{gray!5}
IGPO~\citep{wang2026information}
& Information Gain
& \sgResult{0.414}{0.021}{\sgGoodUp{0.164}}
& \sgResult{0.292}{0.017}{\sgGoodUp{0.093}}
& \sgResult{0.355}{0.016}{\sgGoodUp{0.117}}
& \sgResult{0.137}{0.015}{\sgGoodUp{0.080}}
& \sgResult{0.934}{0.016}{\sgGoodUp{0.004}}
& \sgResult{0.056}{0.012}{\sgGoodDown{0.013}}
& \sgResult{3.028}{0.19}{\sgBadUp{0.813}} \\

Search-E1~\citep{liang2026search}
& Self-distillation
& \sgResult{0.437}{0.020}{\sgGoodUp{0.187}}
& \sgBestResult{0.314}{0.017}{\sgGoodUp{0.115}}
& \sgBestResult{0.372}{0.016}{\sgGoodUp{0.134}}
& \sgResult{0.142}{0.014}{\sgGoodUp{0.085}}
& \sgResult{0.939}{0.015}{\sgGoodUp{0.009}}
& \sgResult{0.054}{0.011}{\sgGoodDown{0.015}}
& \sgResult{2.980}{0.18}{\sgBadUp{0.765}} \\

\rowcolor{gray!5}
SAAS~\citep{tang2026saas}
& Rollout Boundary
& \sgResult{0.428}{0.019}{\sgGoodUp{0.178}}
& \sgResult{0.301}{0.017}{\sgGoodUp{0.102}}
& \sgResult{0.355}{0.016}{\sgGoodUp{0.117}}
& \sgResult{0.138}{0.014}{\sgGoodUp{0.081}}
& \sgResult{0.951}{0.013}{\sgGoodUp{0.021}}
& \sgResult{0.045}{0.010}{\sgGoodDown{0.024}}
& \sgResult{2.122}{0.16}{\sgGoodDown{0.093}} \\

$\beta$-GRPO~\citep{wu2025search}
& Query Confidence
& \sgResult{0.419}{0.020}{\sgGoodUp{0.169}}
& \sgResult{0.295}{0.017}{\sgGoodUp{0.096}}
& \sgResult{0.344}{0.016}{\sgGoodUp{0.106}}
& \sgResult{0.129}{0.014}{\sgGoodUp{0.072}}
& \sgResult{0.948}{0.014}{\sgGoodUp{0.018}}
& \sgResult{0.047}{0.010}{\sgGoodDown{0.022}}
& \sgResult{2.227}{0.15}{\sgBadUp{0.012}} \\

\rowcolor{gray!5}
KbPO~\citep{feng2026trust}
& Certainty+Quality
& \sgResult{0.431}{0.019}{\sgGoodUp{0.181}}
& \sgResult{0.305}{0.016}{\sgGoodUp{0.106}}
& \sgResult{0.358}{0.015}{\sgGoodUp{0.120}}
& \sgResult{0.140}{0.013}{\sgGoodUp{0.083}}
& \sgResult{0.955}{0.012}{\sgGoodUp{0.025}}
& \sgResult{0.043}{0.009}{\sgGoodDown{0.026}}
& \sgResult{2.160}{0.17}{\sgGoodDown{0.055}} \\

\cmidrule(lr){1-9}
\rowcolor{cyan!14}
\textbf{Search-G1 (ours)}
& Representation
& \sgBestResult{0.453}{0.018}{\sgGoodUp{0.203}}
& \sgResult{0.306}{0.016}{\sgGoodUp{0.107}}
& \sgResult{0.368}{0.015}{\sgGoodUp{0.130}}
& \sgBestResult{0.143}{0.013}{\sgGoodUp{0.086}}
& \sgBestResult{0.976}{0.010}{\sgGoodUp{0.046}}
& \sgBestResult{0.034}{0.007}{\sgGoodDown{0.035}}
& \sgBestResult{2.094}{0.14}{\sgGoodDown{0.121}} \\

\midrule
\multicolumn{9}{c}{\textbf{Qwen2.5-7B-Instruct}} \\
\midrule

Base
& None
& 0.212
& 0.203
& 0.250
& 0.041
& n/a
& n/a
& 0.000 \\
\addlinespace[1pt]

\rowcolor{gray!5}
Prompted Search
& Retrieval
& 0.352
& 0.292
& 0.274
& 0.074
& 0.936
& 0.061
& 2.171 \\
\cmidrule(lr){1-9}

Search-R1~\citep{jin2025search}
& Outcome+Direct
& \sgResult{0.417}{0.017}{\sgGoodUp{0.065}}
& \sgResult{0.382}{0.014}{\sgGoodUp{0.090}}
& \sgResult{0.422}{0.013}{\sgGoodUp{0.148}}
& \sgResult{0.154}{0.011}{\sgGoodUp{0.080}}
& \sgResult{0.932}{0.016}{\sgBadDown{0.004}}
& \sgResult{0.046}{0.010}{\sgGoodDown{0.015}}
& \sgResult{3.090}{0.19}{\sgBadUp{0.919}} \\

\rowcolor{gray!5}
GiGPO~\citep{feng2026group}
& State
& \sgResult{0.464}{0.019}{\sgGoodUp{0.112}}
& \sgResult{0.416}{0.015}{\sgGoodUp{0.124}}
& \sgResult{0.436}{0.014}{\sgGoodUp{0.162}}
& \sgResult{0.189}{0.013}{\sgGoodUp{0.115}}
& \sgResult{0.947}{0.014}{\sgGoodUp{0.011}}
& \sgResult{0.043}{0.009}{\sgGoodDown{0.018}}
& \sgResult{2.930}{0.16}{\sgBadUp{0.759}} \\

AgentPRM~\citep{xi2026agentprm}
& Process
& \sgResult{0.455}{0.018}{\sgGoodUp{0.103}}
& \sgResult{0.405}{0.016}{\sgGoodUp{0.113}}
& \sgResult{0.430}{0.015}{\sgGoodUp{0.156}}
& \sgResult{0.205}{0.012}{\sgGoodUp{0.131}}
& \sgResult{0.963}{0.011}{\sgGoodUp{0.027}}
& \sgResult{0.037}{0.008}{\sgGoodDown{0.024}}
& \sgResult{3.080}{0.21}{\sgBadUp{0.909}} \\

\rowcolor{gray!5}
GPT-5 Judge~\citep{singh2025openai}
& External Judge
& \sgResult{0.470}{0.020}{\sgGoodUp{0.118}}
& \sgResult{0.410}{0.016}{\sgGoodUp{0.118}}
& \sgResult{0.440}{0.015}{\sgGoodUp{0.166}}
& \sgResult{0.215}{0.013}{\sgGoodUp{0.141}}
& \sgResult{0.952}{0.014}{\sgGoodUp{0.016}}
& \sgResult{0.033}{0.007}{\sgGoodDown{0.028}}
& \sgResult{2.980}{0.20}{\sgBadUp{0.809}} \\

ARPO~\citep{dong2026agentic}
& Entropy
& \sgResult{0.462}{0.016}{\sgGoodUp{0.110}}
& \sgResult{0.398}{0.014}{\sgGoodUp{0.106}}
& \sgResult{0.420}{0.014}{\sgGoodUp{0.146}}
& \sgResult{0.208}{0.012}{\sgGoodUp{0.134}}
& \sgResult{0.931}{0.015}{\sgBadDown{0.005}}
& \sgResult{0.050}{0.011}{\sgGoodDown{0.011}}
& \sgResult{2.430}{0.14}{\sgBadUp{0.259}} \\

\rowcolor{gray!5}
IGPO~\citep{wang2026information}
& Information Gain
& \sgResult{0.480}{0.019}{\sgGoodUp{0.128}}
& \sgResult{0.425}{0.015}{\sgGoodUp{0.133}}
& \sgResult{0.455}{0.014}{\sgGoodUp{0.181}}
& \sgResult{0.235}{0.014}{\sgGoodUp{0.161}}
& \sgResult{0.942}{0.014}{\sgGoodUp{0.006}}
& \sgResult{0.050}{0.011}{\sgGoodDown{0.011}}
& \sgResult{2.760}{0.17}{\sgBadUp{0.589}} \\

Search-E1~\citep{liang2026search}
& Self-distillation
& \sgResult{0.495}{0.018}{\sgGoodUp{0.143}}
& \sgResult{0.440}{0.015}{\sgGoodUp{0.148}}
& \sgResult{0.465}{0.014}{\sgGoodUp{0.191}}
& \sgResult{0.245}{0.013}{\sgGoodUp{0.171}}
& \sgResult{0.947}{0.014}{\sgGoodUp{0.011}}
& \sgResult{0.048}{0.010}{\sgGoodDown{0.013}}
& \sgResult{2.720}{0.16}{\sgBadUp{0.549}} \\

\rowcolor{gray!5}
SAAS~\citep{tang2026saas}
& Rollout Boundary
& \sgResult{0.486}{0.017}{\sgGoodUp{0.134}}
& \sgResult{0.432}{0.015}{\sgGoodUp{0.140}}
& \sgResult{0.458}{0.014}{\sgGoodUp{0.184}}
& \sgResult{0.239}{0.013}{\sgGoodUp{0.165}}
& \sgResult{0.958}{0.012}{\sgGoodUp{0.022}}
& \sgResult{0.038}{0.008}{\sgGoodDown{0.023}}
& \sgResult{2.050}{0.15}{\sgGoodDown{0.121}} \\

$\beta$-GRPO~\citep{wu2025search}
& Query Confidence
& \sgResult{0.478}{0.018}{\sgGoodUp{0.126}}
& \sgResult{0.424}{0.015}{\sgGoodUp{0.132}}
& \sgResult{0.448}{0.014}{\sgGoodUp{0.174}}
& \sgResult{0.231}{0.013}{\sgGoodUp{0.157}}
& \sgResult{0.953}{0.013}{\sgGoodUp{0.017}}
& \sgResult{0.040}{0.009}{\sgGoodDown{0.021}}
& \sgResult{2.164}{0.14}{\sgGoodDown{0.007}} \\

\rowcolor{gray!5}
KbPO~\citep{feng2026trust}
& Certainty+Quality
& \sgResult{0.490}{0.017}{\sgGoodUp{0.138}}
& \sgResult{0.436}{0.014}{\sgGoodUp{0.144}}
& \sgResult{0.462}{0.014}{\sgGoodUp{0.188}}
& \sgResult{0.242}{0.012}{\sgGoodUp{0.168}}
& \sgResult{0.962}{0.011}{\sgGoodUp{0.026}}
& \sgResult{0.036}{0.008}{\sgGoodDown{0.025}}
& \sgResult{2.106}{0.15}{\sgGoodDown{0.065}} \\

\cmidrule(lr){1-9}
\rowcolor{cyan!14}
\textbf{Search-G1 (ours)}
& Representation
& \sgBestResult{0.512}{0.016}{\sgGoodUp{0.160}}
& \sgBestResult{0.450}{0.014}{\sgGoodUp{0.158}}
& \sgBestResult{0.475}{0.014}{\sgGoodUp{0.201}}
& \sgBestResult{0.254}{0.012}{\sgGoodUp{0.180}}
& \sgBestResult{0.981}{0.009}{\sgGoodUp{0.045}}
& \sgBestResult{0.028}{0.006}{\sgGoodDown{0.033}}
& \sgBestResult{1.860}{0.13}{\sgGoodDown{0.311}} \\

\bottomrule
\end{tabular}
\end{adjustbox}
\end{table}

\paragraph{Baselines.}
We compare with two non-trained references: \textbf{Base}~\citep{ahmed2025qwen}, the closed-book model without retrieval, and \textbf{Prompted Search}~\citep{jin2025search}, which uses the same retrieval interface without policy updates.
The training baselines cover complementary credit mechanisms: \textbf{Search-R1}~\citep{jin2025search} optimizes terminal correctness with a direct-answer bonus in our port; \textbf{GiGPO}~\citep{feng2026group} adds anchor-state groups for stepwise relative credit; \textbf{AgentPRM}~\citep{xi2026agentprm} uses a fixed external process-reward model; our \textbf{GPT-5 Judge}~\citep{singh2025openai} scores completed trajectories with a fixed bespoke evidence-support rubric; \textbf{ARPO}~\citep{dong2026agentic} uses post-tool entropy for adaptive rollouts and attribution; \textbf{IGPO}~\citep{wang2026information} rewards turn-level gains in gold-answer probability; and \textbf{Search-E1}~\citep{liang2026search} combines GRPO with offline self-distillation from sibling contexts.
We additionally compare with three boundary-oriented methods: \textbf{SAAS}~\citep{tang2026saas} contrasts search-disabled and search-enabled rollouts to learn when to search; \textbf{$\beta$-GRPO}~\citep{wu2025search} gates correctness rewards by minimum query-token confidence; and \textbf{KbPO}~\citep{feng2026trust} combines closed-book certainty and retrieval quality into behavior rewards.
All actors share the model, retriever, and interaction/update cap; details and auxiliary costs are in~\Cref{app:implementation_details,tab:baseline_audit}.

\paragraph{Implementation details.}
Trained methods use four independent seeds; the main comparisons, ablations, and sensitivity studies report means across these runs. Fixed-rollout methods sample groups of five trajectories for at most four rounds; adaptive baselines reallocate samples within the same cap. GRPO uses batch size $128$, learning rate $10^{-6}$, and one clipped-update epoch ($\epsilon=0.2$), with actor loss restricted to policy-generated tokens. Every $25$ updates, both readouts are asynchronously refit on a disjoint $300$-question pool, selecting architecture and layer by validation AUC. Reward/refresh settings and final checkpoints are fixed on validation before greedy held-out evaluation. Experiments use eight NVIDIA H20 GPUs (96\,GB) and eight A100 GPUs (80\,GB), with complete settings in~\Cref{app:implementation_details}.

\subsection{Main results}
\label{sec:main}
Search-G1 achieves competitive task accuracy, with higher TC, lower D-Inv, and fewer searches than other search-enabled baselines in the grounding comparisons (\Cref{tab:main,tab:grounding_2wiki}).

\paragraph{Task accuracy across benchmarks.}
As shown in~\Cref{tab:main}, Search-G1 improves mean EM over Prompted Search on every benchmark and at both scales, by $0.086$--$0.203$ at 3B and $0.158$--$0.201$ at 7B. It has the highest 7B mean on all four benchmarks; at 3B, it leads NQ and MuSiQue and remains within $0.008/0.004$ of the best HotpotQA/2Wiki mean. This pattern spans single-hop and multi-hop settings rather than concentrating on one benchmark.

\paragraph{Grounding and search economy.}
The NQ results in~\Cref{tab:main} show the targeted joint behavior: at both scales, Search-G1 has the highest mean EM and TC, the lowest mean D-Inv, and the lowest mean Search/Q among search-enabled methods. Relative to Prompted Search, it raises EM by $0.203/0.160$ while reducing Search/Q by $0.121/0.311$. On 2Wiki, \Cref{tab:grounding_2wiki} shows the same direction: relative to Search-R1, Search-G1 raises mean EM by $0.101/0.053$ and TC by $0.059/0.057$, lowers D-Inv by $0.027/0.024$, and uses $0.590/0.703$ fewer searches at 3B/7B.

\paragraph{Hierarchical inference.}
Across four independent training seeds, hierarchical bootstrap with Holm correction supports improvements over Search-R1 on all four NQ endpoints; EM intervals against the post-hoc metric-wise baseline envelope include zero (\Cref{tab:paired_significance}). These results support competitive accuracy alongside the grounding and search-efficiency gains; \Cref{app:implementation_details} details the statistical protocol.

\begin{figure}[tbp]
\centering
\includegraphics[width=0.7\textwidth]{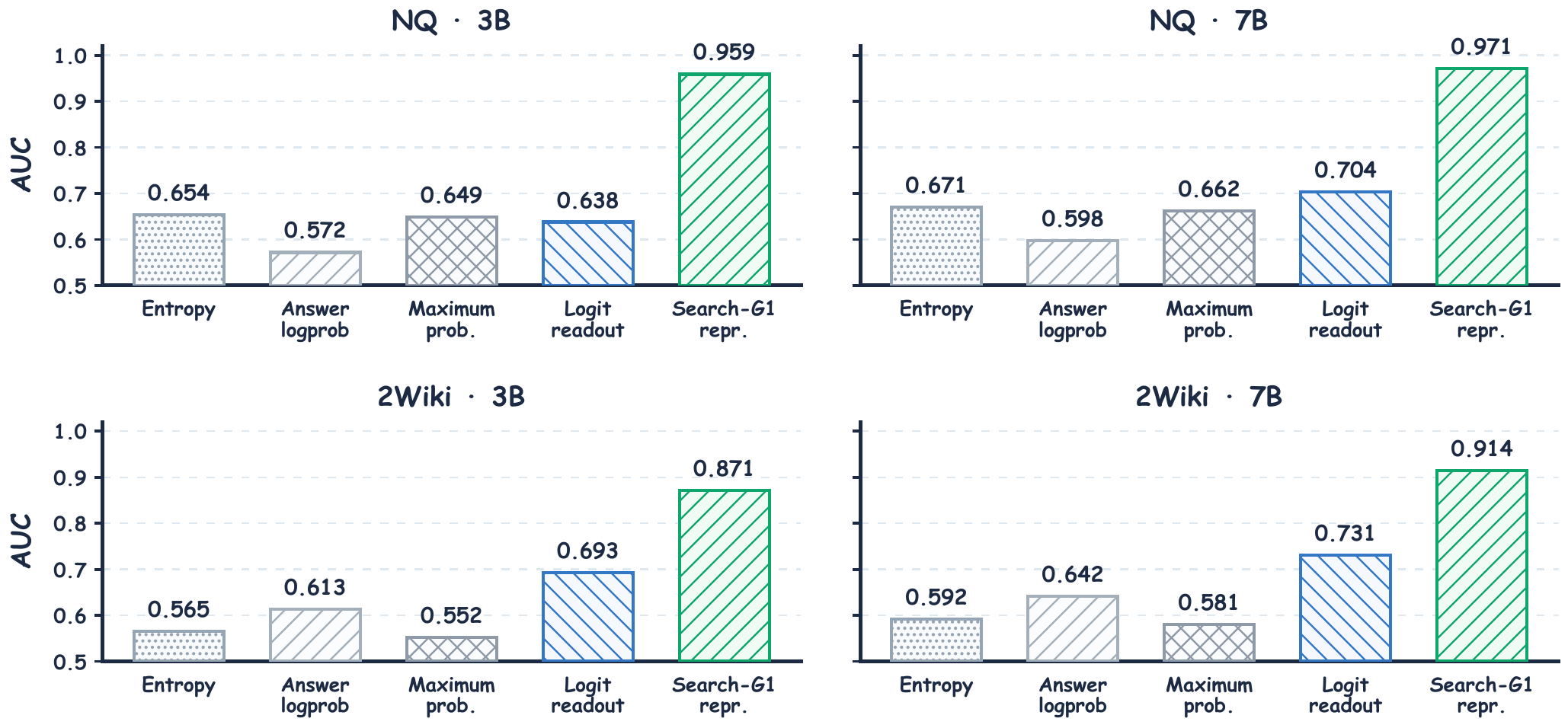}
\caption{Comparison of evidence-reliance predictors AUC on NQ and 2WikiMultiHopQA.}
\label{fig:estimator_fidelity}
\end{figure}

\begin{table}[tbp]
\centering
\setlength{\tabcolsep}{4.5pt}
\renewcommand{\arraystretch}{1.08}
\caption{Comparison of accuracy, grounding, and search cost on 2WikiMultiHopQA. Full-split greedy results.}
\footnotesize
\begin{adjustbox}{max width=\linewidth}
\begin{tabular}{@{}lcccc@{}}
\toprule
Method & EM$\uparrow$ & TC$\uparrow$ & D-Inv$\downarrow$ & Search/Q$\downarrow$ \\
\midrule
\multicolumn{5}{c}{Qwen2.5-3B-Instruct} \\
\midrule
Prompted Search & 0.238 & 0.877 & 0.085 & 2.951 \\
\rowcolor{gray!5}
Search-R1 & 0.267 & 0.913 & 0.058 & 3.402 \\
\rowcolor{cyan!10}
Search-G1 & \textbf{0.368} & \textbf{0.972} & \textbf{0.031} & \textbf{2.812} \\
\midrule
\multicolumn{5}{c}{Qwen2.5-7B-Instruct} \\
\midrule
Prompted Search & 0.274 & 0.887 & 0.079 & 2.892 \\
\rowcolor{gray!5}
Search-R1 & 0.422 & 0.921 & 0.050 & 3.198 \\
\rowcolor{cyan!10}
Search-G1 & \textbf{0.475} & \textbf{0.978} & \textbf{0.026} & \textbf{2.495} \\
\bottomrule
\end{tabular}
\end{adjustbox}
\label{tab:grounding_2wiki}
\end{table}

\begin{table}[tbp]
\centering
\setlength{\tabcolsep}{4pt}
\renewcommand{\arraystretch}{1.08}
\caption{Ablation study of reward components and readout refresh on 2WikiMultiHopQA. Entries report EM/TC/Search/Q.}
\footnotesize
\begin{adjustbox}{max width=\linewidth}
\begin{tabular}{@{}lcccccc@{}}
\toprule
\multirow{2}{*}{Variant} & \multicolumn{3}{c}{Qwen2.5-3B} & \multicolumn{3}{c}{Qwen2.5-7B} \\
\cmidrule(lr){2-4}\cmidrule(lr){5-7}
& EM$\uparrow$ & TC$\uparrow$ & Search/Q$\downarrow$ & EM$\uparrow$ & TC$\uparrow$ & Search/Q$\downarrow$ \\
\midrule
\rowcolor{cyan!10}
Full Search-G1                    & 0.368 & 0.972 & 2.812 & 0.475 & 0.978 & 2.495 \\
No-refresh evaluator              & 0.357 & 0.906 & 3.225 & 0.466 & 0.922 & 2.846 \\
\rowcolor{gray!6}
w/o search cost                   & 0.338 & 0.920 & 3.576 & 0.449 & 0.934 & 3.144 \\
w/o reliance shaping              & 0.354 & 0.797 & 1.992 & 0.463 & 0.829 & 1.798 \\
w/o searched gate                 & 0.347 & 0.877 & 3.119 & 0.444 & 0.897 & 2.756 \\
\rowcolor{gray!6}
w/o direct bonus                  & 0.352 & 0.915 & 2.951 & 0.451 & 0.936 & 2.612 \\
w/o both necessity terms          & 0.339 & 0.861 & 3.286 & 0.436 & 0.881 & 2.905 \\
\rowcolor{gray!6}
Shuffled searched gate            & 0.341 & 0.869 & 3.204 & 0.439 & 0.889 & 2.838 \\
\bottomrule
\end{tabular}
\end{adjustbox}
\label{tab:ablation_component}
\end{table}

\subsection{Analysis}
\label{sec:analysis}

We examine readout fidelity, the alignment of group-relative credit with evidence support, and reliability across policy updates. Behavioral and independent human audits assess the resulting search use and answer support.
\paragraph{Readout and label validity.}
\Cref{fig:estimator_fidelity} shows answer-commit AUC $0.959/0.971$ on NQ and $0.871/0.914$ on 2Wiki. Additionally removing retained reasoning or matching decoding yields $93$--$96\%$ agreement with defaults and positive-rate shifts at most $0.03$ (\Cref{tab:label_validity,tab:label_audit}).

\paragraph{Surface controls.}
In the 3B audit, answer-commit AUC remains $0.934/0.842$ after matching on search count and response length; shuffled labels yield near-chance AUC (\Cref{tab:shortcut_controls}).

\begin{figure}[tbp]
\centering
\includegraphics[width=0.60\linewidth]{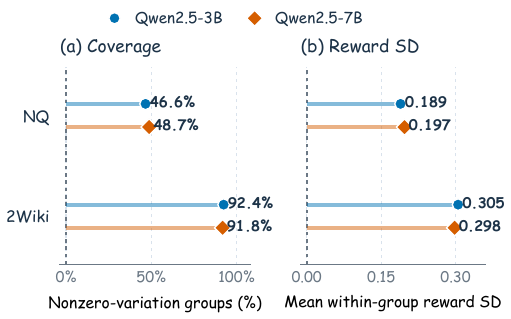}
\caption{Reward variation within same-EM groups. Left: coverage of nonzero-variance groups; right: their mean within-group reward SD.}
\label{fig:reward_variance}
\end{figure}

\paragraph{Credit beyond terminal correctness.}
\Cref{fig:reward_variance} reports nonzero variation in $46.6$--$48.7\%$ of same-EM NQ groups and $91.8$--$92.4\%$ on 2Wiki; in the 3B/2Wiki control, advantage--support correlation is $0.41$ versus $0.02$ after permutation (\Cref{tab:credit_control}). The permutation control tests credit--support association beyond the presence of reward variation.

\begin{figure}[tbp]
\centering
\includegraphics[width=0.78\linewidth]{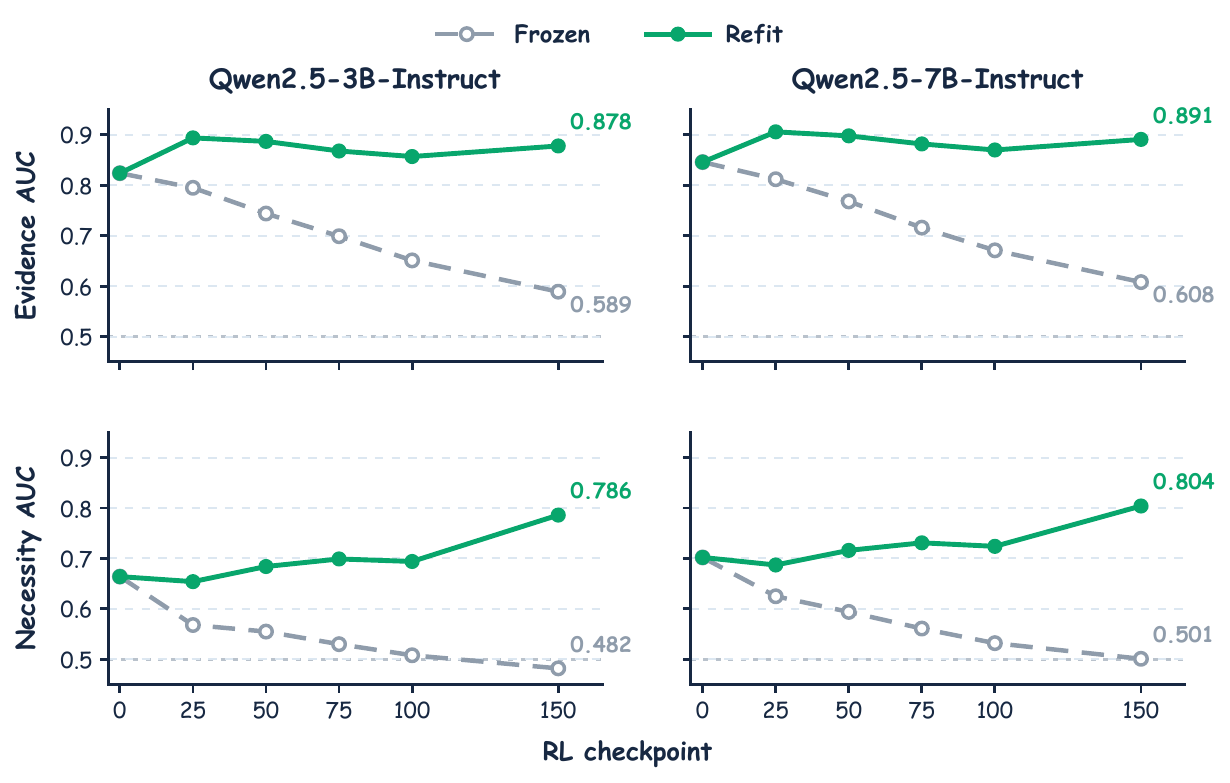}
\caption{Evolution of readout AUC across policy checkpoints on 2WikiMultiHopQA. Top: evidence reliance; bottom: necessity; left: 3B; right: 7B.}
\label{fig:drift}
\end{figure}

\paragraph{Recalibration across policy updates.}
Readouts must remain informative across updates. At checkpoint 150, frozen reliance and necessity AUCs at 3B/7B fall to $0.589/0.608$ and $0.482/0.501$; refitting retains $0.878/0.891$ and $0.786/0.804$ on 2WikiMultiHopQA (\Cref{fig:drift}).

\paragraph{Policy boundary and realized cost.}
On its closed-book-correct slices, Search-G1 has repeated-search rates of $0.040/0.028$. On fixed Base-CB-wrong questions, valid-conditioned EM reaches $0.327/0.356$ with fewer searches than Search-R1 (\Cref{tab:search_efficiency_cb,tab:cbwrong_retrieval}). On NQ, policy-generated and total response-side tokens fall $43$--$44\%$ and $16$--$19\%$ versus Search-R1 (\Cref{tab:cost_audit_nq,tab:resource_accounting}).

\paragraph{Independent grounding.}
Human judgments use neither perturbations nor readouts: Search-G1 reaches $0.842$ Support@Correct on independent correct-searched pools (\Cref{tab:human_grounding}). On $600$ shared NQ IDs, its unconditional grounded accuracy is $0.360\,[0.322,0.399]$, comparable to GPT-5 Judge at $0.355\,[0.317,0.394]$, with the same rounded unsupported rate (\Cref{tab:decomposition}). This common-denominator check complements conditional support; \Cref{tab:cross_dataset_transfer} reports transfer results.

Together, these analyses connect accurate intervention readouts to credit aligned with human-assessed support, while showing that recalibration preserves readout reliability as the policy evolves.

\subsection{Ablations}

\label{sec:ablation}

\paragraph{Component ablations.}
In the four-run means, removing search cost increases Search/Q and lowers EM: more retrieval need not improve accuracy. Removing reliance shaping reduces search but incurs the largest TC loss (\Cref{tab:ablation_component}), while removing refresh lowers EM/TC and increases search.

\paragraph{Necessity factorization.}
As \Cref{prop:gate_normalization} predicts, $n_i$ can change credit direction in heterogeneous groups, but not in all-correct, equal-cost searched groups. Removing the gate \mbox{lowers} EM/TC and raises Search/Q; shuffling $n_i$ also degrades performance (\Cref{app:necessity_factorial}).

\paragraph{Coefficient and refresh sensitivity.}
\Cref{tab:coef_sensitivity,tab:update_sensitivity} show how reward weights and refresh frequency balance objectives: $\lambda_g=0.8$ raises EM at lower TC and more search, $\eta=0.05$ increases Search/Q, and $K=50$ reduces Search/Q but lowers EM and TC.

\paragraph{Unified accounting.}
On $1{,}024$ shared IDs, Search-G1 leads EM ($0.355$), searched-only TC ($0.655$), and EM per marker ($0.225$) at $1.580$ searches per question. Shared IDs align questions, and S-TC excludes direct answers; correct-searched subsets remain policy-dependent (\Cref{tab:efficiency_controlled}).

\subsection{Case Study}
\label{sec:case_study}
\Cref{fig:casestudy} shows three selected 2Wiki entity, bridge, and compositional cases. Search-G1 completes each relation chain with one search; comparison trajectories miss a relation or fail to use retrieved evidence. These cases illustrate relation-chain completion and complement reward diagnostics.

\section{Conclusion}

In this paper, we present Search-G1, which conditions evidence-reliance credit on policy-relative retrieval necessity. Two intervention-calibrated readouts amortize estimates of closed-book sufficiency and answer-stage evidence-deletion sensitivity, supporting a correctness-first reward that favors correct direct answers when knowledge suffices, rewards evidence-sensitive search when needed, and penalizes repeated retrieval. Periodic target regeneration and refitting adapt both signals as the policy evolves, without external judge calls during policy optimization. Across four QA benchmarks and two scales, Search-G1 achieves competitive task accuracy, stronger counterfactual evidence dependence, and fewer searches in the main comparisons. Ablations and independent human-support audits further support the reward design in the evaluated QA settings.

\paragraph{Limitations.}
Our evaluation covers short-form QA with static Wikipedia retrieval and two Qwen scales; training uses verifiable gold answers. Open-ended, live-web, multilingual, multimodal, larger-scale, and alternative-RL settings are natural extensions. Efficiency is measured through search and token use. Reliance captures prefix-conditional deletion sensitivity, which may include distribution-shift effects.

\FloatBarrier

\clearpage
\renewcommand{\bibfont}{\small}
\bibliographystyle{plainnat}
\bibliography{main}

@inproceedings{shang2025agentsquare,
  title={Agentsquare: Automatic llm agent search in modular design space},
  author={Shang, Yu and Li, Yu and Zhao, Keyu and Ma, Likai and Liu, Jiahe and Xu, Fengli and Li, Yong},
  booktitle={International Conference on Learning Representations},
  volume={2025},
  pages={3841--3865},
  year={2025}
}

@inproceedings{yin2026deliberative,
  title={Deliberative Searcher: Improving LLM Reliability via Reinforcement Learning with Constraints},
  author={Yin, Zhenyun and Wang, Shujie and Wang, Xuhong and Ma, Xingjun and Wang, Yingchun},
  booktitle={Proceedings of the 64th Annual Meeting of the Association for Computational Linguistics (Volume 1: Long Papers)},
  pages={4340--4354},
  year={2026}
}

@inproceedings{asai2024self,
  title={Self-rag: Learning to retrieve, generate, and critique through self-reflection},
  author={Asai, Akari and Wu, Zeqiu and Wang, Yizhong and Sil, Avi and Hajishirzi, Hannaneh},
  booktitle={International conference on learning representations},
  volume={2024},
  pages={9112--9141},
  year={2024}
}

@inproceedings{jeong2024adaptive,
  title={Adaptive-rag: Learning to adapt retrieval-augmented large language models through question complexity},
  author={Jeong, Soyeong and Baek, Jinheon and Cho, Sukmin and Hwang, Sung Ju and Park, Jong C},
  booktitle={Proceedings of the 2024 conference of the north american chapter of the association for computational linguistics: Human language technologies (volume 1: Long papers)},
  pages={7036--7050},
  year={2024}
}

@article{zhang2026self,
  title={Self-evolving LLM agents with in-distribution Optimization},
  author={Zhang, Yudi and Fang, Meng and Chen, Zhenfang and Pechenizkiy, Mykola},
  journal={arXiv preprint arXiv:2606.07367},
  year={2026}
}

@inproceedings{zhan2026safesearch,
  title={Safesearch: Do not trade safety for utility in llm search agents},
  author={Zhan, Qiusi and Budiman-Chan, Angeline and Zayed, Abdelrahman and Guo, Xingzhi and Kang, Daniel and Kim, Joo-Kyung},
  booktitle={Findings of the Association for Computational Linguistics: EACL 2026},
  pages={2800--2815},
  year={2026}
}

@article{jin2025search,
  title={Search-r1: Training llms to reason and leverage search engines with reinforcement learning},
  author={Jin, Bowen and Zeng, Hansi and Yue, Zhenrui and Yoon, Jinsung and Arik, Sercan and Wang, Dong and Zamani, Hamed and Han, Jiawei},
  journal={arXiv preprint arXiv:2503.09516},
  year={2025}
}

@article{wang2025spa,
  title={Spa-rl: Reinforcing llm agents via stepwise progress attribution},
  author={Wang, Hanlin and Leong, Chak Tou and Wang, Jiashuo and Wang, Jian and Li, Wenjie},
  journal={arXiv preprint arXiv:2505.20732},
  year={2025}
}

@inproceedings{xi2026agentprm,
  title={Agentprm: Process reward models for llm agents via step-wise promise and progress},
  author={Xi, Zhiheng and Liao, Chenyang and Li, Guanyu and Zhang, Zhihao and Chen, Wenxiang and Wang, Binghai and Jin, Senjie and Zhou, Yuhao and Guan, Jian and Wu, Wei and others},
  booktitle={Proceedings of the ACM Web Conference 2026},
  pages={4184--4195},
  year={2026}
}

@article{liu2026agenthallu,
  title={AgentHallu: Benchmarking automated hallucination attribution of LLM-based agents},
  author={Liu, Xuannan and Yang, Xiao and Li, Zekun and Li, Peipei and He, Ran},
  journal={arXiv preprint arXiv:2601.06818},
  year={2026}
}

@article{kim2026pair,
  title={PAIR: Prefix-Aware Internal Reward Model for Multi-Turn Agent Optimization},
  author={Kim, Wonjoong and In, Yeonjun and Park, Sangwu and Lee, Dongha and Park, Chanyoung},
  journal={arXiv preprint arXiv:2605.17877},
  year={2026}
}

@article{tang2026saas,
  title={SAAS: Self-Aware Reinforcement Learning for Over-Search Mitigation in Agentic Search},
  author={Tang, Yunbo and Yang, Chengyi and Liu, Shiyu and Xiang, Zhishang and Chen, Zerui and Zhang, Qinggang and Su, Jinsong},
  journal={arXiv preprint arXiv:2605.29796},
  year={2026}
}

@inproceedings{feng2026trust,
  title={Trust Within? Seek Beyond? Knowledge Boundary Aware Policy Optimization for Agentic Search},
  author={Feng, Tao and Jiang, Xinke and Hu, Xinyan and Zhang, Yonggang and Tao, Zhen and Zhang, Wentao and Liu, Boyang and Jiang, Wenhao and Wu, Chao},
  booktitle={Proceedings of the 64th Annual Meeting of the Association for Computational Linguistics (Volume 1: Long Papers)},
  pages={27664--27682},
  year={2026}
}

@inproceedings{tuyls2026representation,
  title={Representation-based exploration for language models: From test-time to post-training},
  author={Tuyls, Jens and Foster, Dylan and Krishnamurthy, Akshay and Ash, Jordan},
  booktitle={International Conference on Learning Representations},
  volume={2026},
  pages={120621--120645},
  year={2026}
}

@article{healy2026internal,
  title={Internal representations as indicators of hallucinations in agent tool selection},
  author={Healy, Kait and Srinivasan, Bharathi and Madathil, Visakh and Wu, Jing},
  journal={arXiv preprint arXiv:2601.05214},
  year={2026}
}

@inproceedings{dong2026agentic,
  title={Agentic reinforced policy optimization},
  author={Dong, Guanting and Mao, Hangyu and Ma, Kai and Bao, Licheng and Chen, Yifei and Wang, Zhongyuan and Chen, Zhongxia and Du, Jiazhen and Wang, Huiyang and Zhang, Fuzheng and others},
  booktitle={International Conference on Learning Representations},
  volume={2026},
  pages={16981--17017},
  year={2026}
}

@inproceedings{wang2026information,
  title={Information gain-based policy optimization: A simple and effective approach for multi-turn search agents},
  author={Wang, Guoqing and Dai, Sunhao and Ye, Guangze and Gan, Zeyu and Yao, Wei and Deng, Yong and Wu, Xiaofeng and others},
  booktitle={International Conference on Learning Representations},
  volume={2026},
  pages={104502--104529},
  year={2026}
}

@article{song2025r1,
  title={R1-searcher: Incentivizing the search capability in llms via reinforcement learning},
  author={Song, Huatong and Jiang, Jinhao and Min, Yingqian and Chen, Jie and Chen, Zhipeng and Zhao, Wayne Xin and Fang, Lei and Wen, Ji-Rong},
  journal={arXiv preprint arXiv:2503.05592},
  year={2025}
}

@article{hu2025openreward,
  title={Openreward: Learning to reward long-form agentic tasks via reinforcement learning},
  author={Hu, Ziyou and Shi, Zhengliang and Zhu, Minghang and Li, Haitao and Sun, Teng and Ren, Pengjie and Verberne, Suzan and Ren, Zhaochun},
  journal={arXiv preprint arXiv:2510.24636},
  year={2025}
}

@article{du2019liir,
  title={Liir: Learning individual intrinsic reward in multi-agent reinforcement learning},
  author={Du, Yali and Han, Lei and Fang, Meng and Liu, Ji and Dai, Tianhong and Tao, Dacheng},
  journal={Advances in neural information processing systems},
  volume={32},
  year={2019}
}

@inproceedings{mallen2023not,
  title={When not to trust language models: Investigating effectiveness of parametric and non-parametric memories},
  author={Mallen, Alex and Asai, Akari and Zhong, Victor and Das, Rajarshi and Khashabi, Daniel and Hajishirzi, Hannaneh},
  booktitle={Proceedings of the 61st annual meeting of the association for computational linguistics (volume 1: Long papers)},
  pages={9802--9822},
  year={2023}
}

@article{wang2023survey,
  title={Survey on factuality in large language models: Knowledge, retrieval and domain-specificity},
  author={Wang, Cunxiang and Liu, Xiaoze and Yue, Yuanhao and Tang, Xiangru and Zhang, Tianhang and Jiayang, Cheng and Yao, Yunzhi and Gao, Wenyang and Hu, Xuming and Qi, Zehan and others},
  journal={arXiv preprint arXiv:2310.07521},
  year={2023}
}

@article{gao2025beyond,
  title={Beyond ten turns: Unlocking long-horizon agentic search with large-scale asynchronous rl},
  author={Gao, Jiaxuan and Fu, Wei and Xie, Minyang and Xu, Shusheng and He, Chuyi and Mei, Zhiyu and Zhu, Banghua and Wu, Yi},
  journal={arXiv preprint arXiv:2508.07976},
  year={2025}
}

@inproceedings{wu2025search,
  title={Search Wisely: Mitigating Sub-optimal Agentic Searches By Reducing Uncertainty},
  author={Wu, Peilin and Zhang, Mian and Zhang, Xinlu and Du, Xinya and Chen, Zhiyu},
  booktitle={Proceedings of the 2025 Conference on Empirical Methods in Natural Language Processing},
  pages={19734--19745},
  year={2025}
}

@inproceedings{petroni2019language,
  title={Language models as knowledge bases?},
  author={Petroni, Fabio and Rockt{\"a}schel, Tim and Riedel, Sebastian and Lewis, Patrick and Bakhtin, Anton and Wu, Yuxiang and Miller, Alexander},
  booktitle={Proceedings of the 2019 conference on empirical methods in natural language processing and the 9th international joint conference on natural language processing (EMNLP-IJCNLP)},
  pages={2463--2473},
  year={2019}
}

@inproceedings{geva2021transformer,
  title={Transformer feed-forward layers are key-value memories},
  author={Geva, Mor and Schuster, Roei and Berant, Jonathan and Levy, Omer},
  booktitle={Proceedings of the 2021 conference on empirical methods in natural language processing},
  pages={5484--5495},
  year={2021}
}

@article{kadavath2022language,
  title={Language models (mostly) know what they know},
  author={Kadavath, Saurav and Conerly, Tom and Askell, Amanda and Henighan, Tom and Drain, Dawn and Perez, Ethan and Schiefer, Nicholas and Hatfield-Dodds, Zac and DasSarma, Nova and Tran-Johnson, Eli and others},
  journal={arXiv preprint arXiv:2207.05221},
  year={2022}
}

@inproceedings{azaria2023internal,
  title={The internal state of an LLM knows when it’s lying},
  author={Azaria, Amos and Mitchell, Tom},
  booktitle={Findings of the Association for Computational Linguistics: EMNLP 2023},
  pages={967--976},
  year={2023}
}

@inproceedings{chuang2024dola,
  title={Dola: Decoding by contrasting layers improves factuality in large language models},
  author={Chuang, Yung-Sung and Xie, Yujia and Luo, Hongyin and Kim, Yoon and Glass, James R and He, Pengcheng},
  booktitle={International Conference on Learning Representations},
  volume={2024},
  pages={54158--54183},
  year={2024}
}

@article{ahmed2025qwen,
  title={Qwen 2.5: A comprehensive review of the leading resource-efficient llm with potentioal to surpass all competitors},
  author={Ahmed, Imtiaz and Islam, Sadman and Datta, Partha Protim and Kabir, Imran and Chowdhury, Md Naseef Ur Rahman and Haque, Ahshanul},
  year={2025},
  publisher={TechRxiv}
}

@article{kwiatkowski2019natural,
  title={Natural questions: a benchmark for question answering research},
  author={Kwiatkowski, Tom and Palomaki, Jennimaria and Redfield, Olivia and Collins, Michael and Parikh, Ankur and Alberti, Chris and Epstein, Danielle and Polosukhin, Illia and Devlin, Jacob and Lee, Kenton and others},
  journal={Transactions of the Association for Computational Linguistics},
  volume={7},
  pages={453--466},
  year={2019},
  publisher={MIT Press One Rogers Street, Cambridge, MA 02142-1209, USA journals-info~…}
}

@article{shao2024deepseekmath,
  title={Deepseekmath: Pushing the limits of mathematical reasoning in open language models},
  author={Shao, Zhihong and Wang, Peiyi and Zhu, Qihao and Xu, Runxin and Song, Junxiao and Bi, Xiao and Zhang, Haowei and Zhang, Mingchuan and Li, YK and Wu, Yang and others},
  journal={arXiv preprint arXiv:2402.03300},
  year={2024}
}

@inproceedings{yang2018hotpotqa,
  title={HotpotQA: A dataset for diverse, explainable multi-hop question answering},
  author={Yang, Zhilin and Qi, Peng and Zhang, Saizheng and Bengio, Yoshua and Cohen, William and Salakhutdinov, Ruslan and Manning, Christopher D},
  booktitle={Proceedings of the 2018 conference on empirical methods in natural language processing},
  pages={2369--2380},
  year={2018}
}

@article{zhao2026understanding,
  title={Understanding parametric and contextual knowledge reconciliation within large language models},
  author={Zhao, Jun and Yang, Yongzhuo and Hu, Xiang and Tong, Jingqi and Lu, Yi and Wu, Wei and Gui, Tao and Zhang, Qi and Huang, Xuanjing},
  journal={Advances in Neural Information Processing Systems},
  volume={38},
  pages={102978--103012},
  year={2026}
}

@article{hernandez2023inspecting,
  title={Inspecting and editing knowledge representations in language models},
  author={Hernandez, Evan and Li, Belinda Z and Andreas, Jacob},
  journal={arXiv preprint arXiv:2304.00740},
  year={2023}
}

@inproceedings{zhao2026pre,
  title={Pre-training limited memory language models with internal and external knowledge},
  author={Zhao, Linxi and Zalouk, Sofian and Belardi, Christian and Lovelace, Justin and Zhou, Jin and Noonan, Ryan and Go, Dongyoung and Weinberger, Kilian and Artzi, Yoav and Sun, Jennifer},
  booktitle={International Conference on Learning Representations},
  volume={2026},
  pages={60117--60152},
  year={2026}
}

@inproceedings{ho2020constructing,
  title={Constructing a multi-hop qa dataset for comprehensive evaluation of reasoning steps},
  author={Ho, Xanh and Nguyen, Anh-Khoa Duong and Sugawara, Saku and Aizawa, Akiko},
  booktitle={Proceedings of the 28th International Conference on Computational Linguistics},
  pages={6609--6625},
  year={2020}
}

@article{trivedi2022musique,
  title={musique: Multihop questions via single-hop question composition},
  author={Trivedi, Harsh and Balasubramanian, Niranjan and Khot, Tushar and Sabharwal, Ashish},
  journal={Transactions of the Association for Computational Linguistics},
  volume={10},
  pages={539--554},
  year={2022},
  publisher={MIT Press One Broadway, 12th Floor, Cambridge, Massachusetts 02142, USA~…}
}

@article{feng2026group,
  title={Group-in-group policy optimization for llm agent training},
  author={Feng, Lang and Xue, Zhenghai and Liu, Tingcong and An, Bo},
  journal={Advances in Neural Information Processing Systems},
  volume={38},
  pages={46375--46408},
  year={2026}
}

@article{ma2025pou,
  title={Pou: Proof-of-use to counter tool-call hacking in deepresearch agents},
  author={Ma, SHengjie and Deng, Chenlong and Mao, Jiaxin and Huang, Jiadeng and Wang, Teng and Wu, Junjie and Zhang, Changwang and others},
  journal={arXiv e-prints},
  pages={arXiv--2510},
  year={2025}
}

@article{singh2025openai,
  title={Openai gpt-5 system card},
  author={Singh, Aaditya and Fry, Adam and Perelman, Adam and Tart, Adam and Ganesh, Adi and El-Kishky, Ahmed and McLaughlin, Aidan and Low, Aiden and Ostrow, AJ and Ananthram, Akhila and others},
  journal={arXiv preprint arXiv:2601.03267},
  year={2025}
}

@article{liang2026search,
  title={Search-E1: Self-Distillation Drives Self-Evolution in Search-Augmented Reasoning},
  author={Liang, Zihan and Ma, Yufei and Chen, Ben and Qian, Zhipeng and Zhang, Xuxin and Dai, Huangyu and Mao, Lingtao},
  journal={arXiv preprint arXiv:2605.22511},
  year={2026}
}

@article{cheng2026adversarial,
  title={Adversarial Orthogonal Disentanglement for LVLM Hallucination Mitigation},
  author={Cheng, Ruoxi and Ma, Haoxuan and Hai, Zhengfei and Huang, Yiyan and Duan, Ranjie and Zhang, Tianle and Yang, Xu and Ye, Ziyi and Ma, Xingjun},
  journal={arXiv preprint arXiv:2605.25377},
  year={2026}
}

@inproceedings{zhang2026criticsearch,
  title={Criticsearch: Fine-grained credit assignment for search agents via a retrospective critic},
  author={Zhang, Yaocheng and Huang, Haohuan and Song, Zijun and Zhao, Zijie and Zhang, Qichao and Zhu, Yuanheng and Zhao, Dongbin},
  booktitle={Findings of the Association for Computational Linguistics: ACL 2026},
  pages={12272--12290},
  year={2026}
}

@inproceedings{sun2025redeep,
  title={Redeep: Detecting hallucination in retrieval-augmented generation via mechanistic interpretability},
  author={Sun, Zhongxiang and Zang, Xiaoxue and Zheng, Kai and Xu, Jun and Zhang, Xiao and Yu, Weijie and Song, Yang and Li, Han},
  booktitle={International Conference on Learning Representations},
  volume={2025},
  pages={50250--50279},
  year={2025}
}

@inproceedings{ferrando2025know,
  title={Do i know this entity? knowledge awareness and hallucinations in language models},
  author={Ferrando, Javier and Obeso, Oscar and Rajamanoharan, Senthooran and Nanda, Neel},
  booktitle={International Conference on Learning Representations},
  volume={2025},
  pages={77965--78000},
  year={2025}
}

@inproceedings{qi2024model,
  title={Model internals-based answer attribution for trustworthy retrieval-augmented generation},
  author={Qi, Jirui and Sarti, Gabriele and Fern{\'a}ndez, Raquel and Bisazza, Arianna},
  booktitle={Proceedings of the 2024 Conference on Empirical Methods in Natural Language Processing},
  pages={6037--6053},
  year={2024}
}

@article{wang2026co,
  title={Co-Evolution of Policy and Internal Reward for Language Agents},
  author={Wang, Xinyu and Wu, Hanwei and Song, Jingwei and Zhang, Shuyuan and Zhang, Jiayi and Kong, Fanqi and Kwok, Tung Sum Thomas and Chang, Xiao-Wen and Luo, Yuyu and Wu, Chenglin and others},
  journal={arXiv preprint arXiv:2604.03098},
  year={2026}
}

@article{chen2026efficient,
  title={Efficient Agentic Reinforcement Learning with On-Policy Intrinsic Knowledge Boundary Enhancement},
  author={Chen, Dingwei and Zong, Zefang and Ma, Zhipeng and Luo, Leo and Li, Yang and Li, Chengming and Chen, Peng and Jiang, Jie},
  journal={arXiv preprint arXiv:2605.26952},
  year={2026}
}

@article{tan2026ctrl,
  title={CTRL-RAG: Contrastive Likelihood Reward Based Reinforcement Learning for Context-Faithful RAG Models},
  author={Tan, Zhehao and Jiao, Yihan and Yang, Dan and Wang, Junjie and Sun, Duolin and Feng, Jie and Wang, Xidong and Liu, Lei and Shen, Yue and Wang, Jian and others},
  journal={arXiv preprint arXiv:2603.04406},
  year={2026}
}

@article{guo2026contextual,
  title={Contextual Information Policy Optimization for Search Agents},
  author={Guo, Xingyu and Chen, Wei and Yang, Linlin and Zhang, Baochang},
  journal={arXiv preprint arXiv:2608.06128},
  year={2026}
}

\clearpage
\appendix
\section{Appendix}

\label{sec:appendix}

\setcounter{table}{0}
\setcounter{figure}{0}
\renewcommand{\thetable}{A\arabic{table}}
\renewcommand{\thefigure}{A\arabic{figure}}

\subsection{Methodology supplements}
\label{app:methodology_supp}

\paragraph{Token likelihood ratio.}
For a selected token $w_{i,j,t}$, the current-to-rollout-policy likelihood ratio in~\Cref{eq:masked_grpo} is
\begin{equation}
\label{eq:policy_ratio}
\rho_{i,j,t}
=
\frac{
\pi_\theta(w_{i,j,t}\mid\tau_{i,j,<t})
}{
\pi_{\mathrm{old}}(w_{i,j,t}\mid\tau_{i,j,<t})
}.
\end{equation}
\paragraph{Distributed implementation of the masked objective.}
For the masked average in~\Cref{eq:masked_grpo}, let $\mathcal B_{s,w}$ denote the rank-local micro-batch on data-parallel worker $w$ at gradient-accumulation step $s$. With $W$ workers and $S$ accumulation steps, and assuming every local micro-batch contains at least one selected token, the implementation-level reduction is
\begin{equation}
\label{eq:masked_average}
\left\langle f_{i,j,t}\right\rangle_{\mathcal{M},\mathcal B}
=
\frac{1}{SW}
\sum_{s=1}^{S}\sum_{w=1}^{W}
\frac{
\sum_{(i,j)\in\mathcal B_{s,w}}\sum_t M_{i,j,t}f_{i,j,t}}
{
\sum_{(i,j)\in\mathcal B_{s,w}}\sum_t M_{i,j,t}}.
\end{equation}
This weights rank-local micro-batches equally, with per-token weights inversely proportional to each micro-batch's selected-token count rather than uniform globally.

\paragraph{Clipped sampled-token KL.}
For a generated token $w_{i,j,t}$, the implemented low-variance penalty is
\begin{equation}
\label{eq:clipped_lv_kl}
\begin{aligned}
u_{i,j,t}
&=\log\pi_{\mathrm{ref}}(w_{i,j,t}\mid\tau_{i,j,<t})
-\log\pi_\theta(w_{i,j,t}\mid\tau_{i,j,<t}),
\\
\widehat{\mathrm{KL}}^{\mathrm{LV}}_{i,j,t}
&=\operatorname{clip}\!\left(
\exp(u_{i,j,t})-u_{i,j,t}-1,
-\kappa,\kappa
\right).
\end{aligned}
\end{equation}
Since $\exp(u)-u-1\ge0$, the negative clipping bound is inactive; only the upper bound limits outlier token penalties and, when active, truncates their gradients.

\paragraph{Necessity-factorization definitions.}
\label{app:necessity_factorial}
The searched branch uses $n_i d_{i,j}$ and the direct branch uses $b_i$. The \emph{w/o searched gate} variant replaces $n_i d_{i,j}$ by $d_{i,j}$; \emph{w/o direct bonus} removes $b_i$; and \emph{w/o both necessity terms} applies both changes. The shuffled control replaces only $n_i$ by a cross-question permutation $\tilde n_i$, redrawn per seed, while keeping $b_i$ paired with its original question. Thus the shuffle tests searched-branch assignment rather than the complete necessity signal or a uniform increase in reward magnitude; outcomes are reported in~\Cref{tab:ablation_component}. On Qwen2.5-3B, the shuffled control ($0.341/0.869/3.204$) lands close to \emph{w/o both necessity terms} ($0.339/0.861/3.286$) and below the full method ($0.368/0.972/2.812$). This ordering is consistent with value from the within-question pairing of $n_i$ and $d_{i,j}$ rather than merely from the presence of a necessity term.

\paragraph{Proof and degenerate cases for \Cref{prop:gate_normalization}.}
\label{app:gate_diag}
We restate the controlled-gate construction of~\Cref{sec:method_optimization} in full. For a fixed question $i$, $\gamma\in[0,1]$ replaces $n_i$ only in the correct-searched reliance term, while $b_i$ and all other reward branches remain fixed; $\gamma=n_i$ recovers Search-G1 and $\gamma=1$ gives the controlled no-gate comparison, which changes only the searched reward coefficient and does not redefine $b_i$. Define
\begin{equation}
\label{eq:gate_reward_decomposition}
\begin{aligned}
x_{i,j}
&=
\begin{cases}
d_{i,j}, & v_{i,j}=1,\ e_{i,j}=1,\ N_{i,j}>0,\\
0, & \text{otherwise},
\end{cases}
\\
\mathbf r_i(\gamma)
&=
\mathbf a_i+\lambda_g\gamma\mathbf x_i,
\qquad
\mathbf a_i:=\mathbf r_i(0).
\end{aligned}
\end{equation}

Let
\[
H=I_G-\frac{1}{G}\mathbf{1}\mathbf{1}^\top,
\qquad
\mathbf u_i=H\mathbf a_i,
\qquad
\mathbf v_i=H\mathbf x_i,
\]
and write $\mathbf z_i(\gamma)=\mathbf u_i+\lambda_g\gamma\mathbf v_i$. Since $\sigma_i(\gamma)$ is the sample standard deviation,
\[
\sigma_i(\gamma)
=
\frac{\|\mathbf z_i(\gamma)\|_2}{\sqrt{G-1}}.
\]

\begin{proof}[Proof of \Cref{prop:gate_normalization}]
Centering~\Cref{eq:gate_reward_decomposition} gives
$H\mathbf r_i(\gamma)=\mathbf u_i+\lambda_g\gamma\mathbf v_i=\mathbf z_i(\gamma)$. Substituting the sample standard deviation gives~\Cref{eq:gated_advantage}; its $j$th component is the scalar advantage in~\Cref{eq:group_advantage}. For $\delta=0$,
\begin{equation}
\label{eq:gated_advantage_direction}
\mathbf A_i^0(\gamma)
=
\sqrt{G-1}\,
\frac{\mathbf z_i(\gamma)}{\|\mathbf z_i(\gamma)\|_2}.
\end{equation}
Thus normalization removes centered reward scale and retains only direction. Suppose $\mathbf u_i$ and $\mathbf v_i$ are linearly independent but the centered vectors for distinct $\gamma_1$ and $\gamma_2$ have the same direction. Then for some $q>0$,
\[
\mathbf u_i+\lambda_g\gamma_1\mathbf v_i
=
q(\mathbf u_i+\lambda_g\gamma_2\mathbf v_i).
\]
Linear independence forces $q=1$ and then $\gamma_1=\gamma_2$, a contradiction. If $\mathbf u_i$ and $\mathbf v_i$ are collinear, all nonzero $\mathbf z_i(\gamma)$ are collinear, and any global reversal between two gate values must pass through $\mathbf z_i(\gamma)=\mathbf 0$. The denominator in~\Cref{eq:gated_advantage} is positive for $\delta>0$, so it changes magnitude but not direction; consequently $\operatorname{sign}(A_{i,j}^\delta(\gamma))=\operatorname{sign}(z_{i,j}(\gamma))$, and the credit of trajectory $j$ changes sign between two gate values exactly when $z_{i,j}(\gamma_1)z_{i,j}(\gamma_2)<0$.

For the strict homogeneous case, suppose every trajectory is valid, correct, searched, and has the same search cost $c_i$. Then $\mathbf a_i=(1-\eta c_i)\mathbf 1$, so $\mathbf u_i=\mathbf 0$ and $\mathbf v_i=H\mathbf d_i$. Let $s_{d,i}$ be the sample standard deviation of $\{d_{i,j}\}_{j=1}^G$. \Cref{eq:gated_advantage} reduces componentwise to
\begin{equation}
\label{eq:homogeneous_gated_advantage}
A_{i,j}^\delta(\gamma)
=
\frac{
\lambda_g\gamma(d_{i,j}-\bar d_i)
}{
\lambda_g\gamma s_{d,i}+\delta
}.
\end{equation}
For $\gamma>0$ and $s_{d,i}>0$,
\begin{equation}
\label{eq:homogeneous_gate_scaling}
\mathbf A_i^0(\gamma)
=
\frac{H\mathbf d_i}{s_{d,i}},
\qquad
\mathbf A_i^\delta(\gamma)
=
\frac{\lambda_g\gamma s_{d,i}}
{\lambda_g\gamma s_{d,i}+\delta}
\mathbf A_i^0(\gamma).
\end{equation}
Hence every positive gate preserves rollout ranking, advantage signs, and credit direction; it cancels from the advantage values when $\delta=0$ and only rescales the entire vector when $\delta>0$.
\end{proof}

\paragraph{Boundary cases.}
If $\gamma=0$, the stabilized advantage in the strict homogeneous case is $\mathbf A_i^\delta(0)=\mathbf 0$, whereas the idealized $\delta=0$ expression is $0/0$ and undefined. If $s_{d,i}=0$, reliance provides no within-group tie-breaking; if $\lambda_g=0$, the gate has no effect by construction. A zero vector $\mathbf z_i(\gamma)$ is likewise degenerate.

\paragraph{Mixed groups.}
Outcome branches, direct-versus-search branches, and unequal search costs can make $\mathbf u_i\neq\mathbf{0}$, so the gate reweights $\mathbf u_i$ against $\mathbf v_i$ in~\Cref{eq:gated_advantage}. Non-collinearity of these centered components, rather than heterogeneous composition alone, is what makes the normalized credit direction gate-dependent. For any two nonzero comparison vectors, the cosine diagnostic is exactly the cosine between $\mathbf z_i(\gamma_1)$ and $\mathbf z_i(\gamma_2)$, because the stabilizer changes only their positive scalar magnitudes.

Cosine similarity is undefined and sign-flip statistics are convention-dependent for zero comparison vectors. We therefore rely on the analytic characterization above; the controlled gate ablation in~\Cref{tab:ablation_component} provides complementary downstream evidence.

\subsection{Experimental setup and reproduction}
\label{app:setup_supp}

\subsubsection{Implementation details}
\label{app:implementation_details}

\paragraph{Data and retrieval.}
Unless stated otherwise, methods share policy-training data, the retriever, and the global interaction/update cap. Benchmark examples come from \texttt{RUC-NLPIR/FlashRAG\_datasets} using the benchmark-native training and available held-out evaluation splits. Retrieval uses the Wikipedia 2018 corpus (\texttt{wiki-18.jsonl}), \texttt{intfloat/e5-base-v2} mean-pooled embeddings, and a FAISS Flat inner-product index, returning three passages per executed retrieval action.

\paragraph{Rollout and optimization.}
We allow four interaction rounds and maximum prompt/\allowbreak response/\allowbreak start/\allowbreak observation lengths of 4096/500/2048/500. Fixed-rollout methods use group size five; adaptive methods reallocate samples within the same cap. GRPO uses sample-standard-deviation normalization with $\delta=10^{-6}$, temperature 1.0, top-$p=0.95$, top-$k=-1$, batch size 128, learning rate $10^{-6}$, and warmup ratio 0.285.

\paragraph{Policy update and evaluation.}
The clipped update uses one epoch, $\epsilon=0.2$, mini-/micro-batches of 32/8, entropy coefficient $\alpha_{\mathrm H}=0.001$, and the sampled-token KL term in~\Cref{eq:clipped_lv_kl} with $\beta_{\mathrm{KL}}=0.001$ and fixed $\kappa=10$. Prompt, observation, and padding tokens are excluded from the actor loss. Unless a table specifies sampling or a fixed subset, single-trajectory metrics use greedy decoding on the full benchmark-specific held-out split. Each evaluation checkpoint is fixed before this final evaluation; held-out test metrics are never used to select it.

\paragraph{Data separation.}
Policy-training questions, the 300-question readout-calibration pool, validation/architecture-selection groups, and final benchmark-evaluation questions are disjoint and identified by stable item IDs. Trained methods use four independently seeded policies, whereas Base and Prompted Search are untrained point estimates.

\paragraph{Statistical unit.}
The independent training run is the unit of replication for trained-method comparisons. With four runs, a two-sided 95\% $t$ interval has half-width approximately $1.59\,\mathrm{SD}$, so small mean differences cannot support superiority by mean ordering alone. Using the seed-by-question matrices, we propagate both sources of variation. Accordingly, \Cref{tab:paired_significance} resamples training seeds and, within each sampled seed, paired questions, then applies Holm correction across EM, TC, D-Inv, and Search/Q within each comparator--scale family. Search-R1 is the primary comparator, against which all four endpoints differ at both scales. The metric-wise upper-envelope rows select the strongest trained comparator by test mean separately for each endpoint and scale; they are an exploratory secondary comparison. Holm correction covers the four endpoints within each comparator--scale family, not this post-hoc comparator selection. Question resampling does not increase the four independent training replications, so the resulting small $p$ values should not be read as precise evidence of superiority across training runs. Only TC separates at both scales, D-Inv and Search/Q separate at 7B, and the secondary analysis does not establish EM superiority over the upper envelope at either scale.

\begin{table}[tbp]
\centering
\caption{Pairwise performance differences on NQ across training seeds and evaluation questions. Entries report hierarchical-bootstrap 95\% CIs and Holm-adjusted $p$ values. $\Delta$EM/$\Delta$TC are Search-G1 minus comparator; D-Inv reverses this sign; negative $\Delta$Search/Q favors Search-G1; $\dagger$: adjusted $p<0.05$.}
\label{tab:paired_significance}
\setlength{\tabcolsep}{3.3pt}
\renewcommand{\arraystretch}{1.06}
\footnotesize
\begin{adjustbox}{max width=\linewidth}
\begin{tabular}{llcccc}
\toprule
Comparator & Scale & $\Delta$EM & $\Delta$TC & D-Inv benefit & $\Delta$Search/Q \\
\midrule
\multirow{2}{*}{Search-R1}
& 3B & \makecell{$+0.101$\\$[+0.071,+0.131]$\\$.004^{\dagger}$} & \makecell{$+0.059$\\$[+0.038,+0.080]$\\$.004^{\dagger}$} & \makecell{$+0.015$\\$[+0.005,+0.025]$\\$.012^{\dagger}$} & \makecell{$-1.187$\\$[-1.440,-0.934]$\\$.004^{\dagger}$} \\
& 7B & \makecell{$+0.095$\\$[+0.020,+0.170]$\\$.032^{\dagger}$} & \makecell{$+0.049$\\$[+0.030,+0.068]$\\$.004^{\dagger}$} & \makecell{$+0.018$\\$[+0.008,+0.028]$\\$.008^{\dagger}$} & \makecell{$-1.230$\\$[-1.460,-1.000]$\\$.004^{\dagger}$} \\
\midrule
\multirow{2}{*}{Strongest per-metric}
& 3B & \makecell{$+0.016$\\$[-0.012,+0.044]$\\$.320$} & \makecell{$+0.018$\\$[+0.004,+0.032]$\\$.032^{\dagger}$} & \makecell{$+0.004$\\$[-0.003,+0.011]$\\$.320$} & \makecell{$-0.028$\\$[-0.114,+0.058]$\\$.520$} \\
& 7B & \makecell{$+0.017$\\$[-0.006,+0.040]$\\$.180$} & \makecell{$+0.018$\\$[+0.006,+0.030]$\\$.016^{\dagger}$} & \makecell{$+0.005$\\$[+0.001,+0.009]$\\$.048^{\dagger}$} & \makecell{$-0.190$\\$[-0.286,-0.094]$\\$.008^{\dagger}$} \\
\bottomrule
\end{tabular}
\end{adjustbox}
\end{table}

\paragraph{Reward and parser settings.}
We set $\lambda_g=0.6$, $\eta=0.02$, $\alpha_{\mathrm{cb}}=0.5$, $r_{\mathrm{inv}}=-0.1$, and $r_{\mathrm{wrong}}=0.05$, satisfying $r_{\mathrm{inv}}<r_{\mathrm{wrong}}<1-\eta$. The default $(\lambda_g,\eta)=(0.6,0.02)$ is fixed from the group-split calibration/validation data, using the coefficient sweep in~\Cref{tab:coef_sensitivity} and the refresh study in~\Cref{tab:update_sensitivity}; test splits are not used for selection. Baseline-specific coefficients are tuned under the same validation protocol and shared budget. The search cost follows~\Cref{eq:search_cost} and is computed from the decoded policy-generated response only:
$c_{i,j}=\mathbf 1[N_{i,j}>1]=\min\{1,(N_{i,j}-1)_+\}$. Markers are counted on the policy-generated response only, so in-context demonstrations in the prompt are never counted. The same response-only parser is applied to direct, one-search, repeated-search, invalid-action, wrong-answer, and correct-answer trajectories.

\paragraph{Evaluation-time resource accounting.}
The evaluator records response-side marker count, policy-generated tokens, retrieved-observation tokens, and wall-clock time for every question. Policy-generated tokens are obtained from the response-side \texttt{info\_mask}; observation tokens are the response-side attention-mask count minus that quantity. \Cref{tab:cost_audit_nq} reports the 3B per-question columns, and \Cref{tab:resource_accounting} reports the corresponding 7B audit.

\paragraph{Hidden-state extraction.}
Both readouts use the selected decoder-block output (hook on \texttt{layers[L-1]} for layer $L$), before the model's final normalization and without token pooling. The input for $h_{p,i}$ is the initial search-enabled prompt; the input for $h_{\mathrm{ans},i,j}$ appends the response up to, but excluding, \texttt{</answer>}, including the generated answer text. Inputs are tokenized without added special tokens, left-padded, and truncated to $4{,}096$ tokens; each state is taken at the last retained token.

\paragraph{Readout refresh and overhead.}
Readouts are selected from logistic regression and MLPs with hidden sizes 64, 128, and $(256,128)$ on frozen states. At each refit, architecture and layer $\{21,24,27\}$ are selected by group-split validation AUC separately for $D_\phi$ and $B_\psi$. The nominal schedule emits checkpoints every $K=25$ updates and refits both readouts asynchronously on $M_{\mathrm{cal}}=300$ calibration questions. A new evaluator is installed atomically only after both readouts finish; otherwise the previous complete evaluator remains active, so the realized refresh interval may exceed the nominal cadence. Sensitivity appears in~\Cref{tab:update_sensitivity}; logistic candidates use class balancing and MLPs use early stopping with $L_2=10^{-4}$.

Beyond the search-enabled calibration rollouts, target regeneration uses one no-retrieval decode per calibration question and one evidence-deleted decode per valid searched calibration rollout. This intervention-specific count is $M_{\mathrm{cal}}+N_{\mathrm{vs}}\le M_{\mathrm{cal}}(1+G)$: at $M_{\mathrm{cal}}=300$ and $G=5$, at most $1.8$k decodes, or approximately $11$--$12\%$ of the $16{,}000$ rollout decodes in a nominal $K=25$ cycle. Reward-time scoring additionally requires a batched, no-gradient teacher-forced pass through the frozen snapshot, but no LLM-judge call during policy updates.

Scoring rollouts with the hard labels instead of the readouts would change this accounting. Rewarding $z_i$ directly needs one closed-book decode per training question, or $128\times25=3{,}200$ decodes per cycle ($20\%$ of rollout decodes); rewarding $s^{\mathrm{chg}}_{i,j}$ directly needs one evidence-deleted decode per searched rollout, or up to $16{,}000$ decodes ($100\%$), since every scored trajectory lies outside the calibration pool. The readouts therefore replace a per-rollout counterfactual decode with a shared forward pass, and return a graded score at the moment credit is assigned. \Cref{fig:estimator_fidelity,tab:necessity_score_drift} report how closely they track the same targets.

\paragraph{Baseline-specific operations.}
The common on-policy phase uses the same policy-training data, base model, retrieval environment, trajectory format, and global interaction/update cap. Adaptive methods may reallocate rollouts within that cap, while each baseline retains its defining auxiliary computation; the comparison therefore matches shared on-policy exposure rather than end-to-end compute.

GiGPO adds anchor-state rollout groups for stepwise relative credit; AgentPRM uses a fixed external process-reward model; GPT-5 Judge scores completed trajectories with our fixed bespoke support rubric; ARPO uses post-tool entropy to trigger branch sampling and applies shared-prefix advantage attribution; IGPO rewards turn-level gains in gold-answer probability; and Search-E1 adds offline token-level forward-KL distillation from privileged sibling contexts. SAAS retains disabled/enabled rollout contrasts and boundary-aware rewards under an adapted staged schedule; $\beta$-GRPO gates correctness by minimum query-token confidence; and KbPO combines no-retrieval certainty sampling with retrieval-quality and behavior classification to form quadrant rewards. Exact auxiliary computation and shared-environment adaptations are summarized in~\Cref{tab:baseline_audit}.

Our adapted Search-R1 reference augments the original terminal-correctness reward with a fixed $0.5$ correct-direct bonus and uses no necessity or reliance readout. Matching the maximum direct-answer bonus avoids comparison only against a bonus-free objective, but does not make the rewards identical: Search-G1 uses the state-dependent bonus $0.5b_i$, necessity-conditioned reliance credit, and repeated-search cost. We therefore treat Search-R1 as an outcome-plus-directness reference, not an isolated causal estimate of the readouts' effect.

\paragraph{Per-baseline implementation audit.}
\Cref{tab:baseline_audit} records whether a comparator is algorithm-faithful under the shared protocol (F), an adapted port retaining the published core mechanism (A), or a bespoke comparator for this study (B). These codes describe implementation provenance rather than source-code identity.

\paragraph{Bespoke external-judge comparator.}
The \textbf{GPT-5 Judge} is not a reproduction of AgentHallu. GPT-5 scores completed trajectories with a fixed evidence-support rubric and output parser designed for this study~\citep{singh2025openai}. AgentHallu motivates agent-trajectory hallucination assessment but does not define this reward; the judge version and prompt remain fixed across runs~\citep{liu2026agenthallu}.

\begin{table}[tbp]
\centering
\caption{Baseline implementations and auxiliary computation under the shared protocol. ``F'' denotes an algorithm-faithful implementation under the shared protocol, ``A'' an adapted port retaining the published core mechanism, and ``B'' a bespoke comparator.}
\label{tab:baseline_audit}
\setlength{\tabcolsep}{3.5pt}
\renewcommand{\arraystretch}{1.1}
\footnotesize
\begin{tabularx}{\linewidth}{@{}l c >{\raggedright\arraybackslash}X >{\raggedright\arraybackslash}X@{}}
\toprule
Method & Status & Auxiliary compute & Shared-protocol adaptation / provenance \\
\midrule
Search-R1  & A & none                     & outcome reward augmented with fixed direct bonus \\
GiGPO      & F & anchor-state rollout groups & anchor-group size set to shared cap \\
AgentPRM   & A & fixed process-reward-model scoring & checkpoint and prompt fixed \\
GPT-5 Judge& B & fixed external-judge calls & bespoke support rubric; model/prompt fixed \\
ARPO       & F & post-tool entropy + branch rollouts & adaptive allocation within shared cap \\
IGPO       & F & shared-model likelihood passes & confidence estimator from shared model \\
Search-E1  & F & offline sibling-context distillation & distillation buffer within cap \\
SAAS       & A & disabled/enabled rollout groups & boundary reward retained; schedule aligned to cap \\
$\beta$-GRPO & A & query-token log probabilities & confidence gate retained under the shared protocol \\
KbPO       & A & certainty samples + retrieval/behavior judging & quadrant reward ported to shared training data \\
\bottomrule
\end{tabularx}
\end{table}

\subsection{Results analysis and additional diagnostics}
\label{app:results_supp}

\subsubsection{Evaluation definitions and diagnostics}
\label{app:analysis}

\begin{table}[tbp]
\centering
\caption{Dataset structure and closed-book sufficiency across four QA benchmarks for Base Qwen2.5-3B-Instruct. Cells report count (percentage) on the full held-out splits. Structural labels are dataset-provided and are not interchangeable with policy-relative retrieval necessity.}
\label{tab:boundary_composition}
\setlength{\tabcolsep}{3.2pt}
\renewcommand{\arraystretch}{1.08}
\footnotesize
\begin{adjustbox}{max width=\linewidth}
\begin{tabular}{@{}l r r r r r@{}}
\toprule
Dataset & $N$ & Exact CB-correct & Exact CB-wrong & Incl. sufficient & Incl. necessary \\
\midrule
NQ & 3,610 & 512 (14.18\%) & 3,098 (85.82\%) & 761 (21.08\%) & 2,849 (78.92\%) \\
HotpotQA & 7,405 & 1,134 (15.31\%) & 6,271 (84.69\%) & 1,437 (19.41\%) & 5,968 (80.59\%) \\
2Wiki & 12,576 & 2,665 (21.19\%) & 9,911 (78.81\%) & 3,115 (24.77\%) & 9,461 (75.23\%) \\
MuSiQue & 2,417 & 71 (2.94\%) & 2,346 (97.06\%) & 101 (4.18\%) & 2,316 (95.82\%) \\
\bottomrule
\end{tabular}
\end{adjustbox}

\par\vspace{6pt}
\begin{tabularx}{\linewidth}{@{}l >{\raggedright\arraybackslash}X@{}}
\toprule
Dataset & Dataset structure \\
\midrule
NQ & single-hop benchmark; no item-level hop field \\
HotpotQA & 79.92\% bridge; 20.08\% comparison \\
2Wiki & 41.63\% compositional; 24.17\% comparison; 12.32\% inference; 21.88\% bridge--comparison \\
MuSiQue & 51.80\% 2-hop; 31.44\% 3-hop; 16.76\% 4-hop \\
\bottomrule
\end{tabularx}
\end{table}

\begin{table}[tbp]
\centering
\caption{Comparison of search and token costs on NQ with Qwen2.5-3B-Instruct. Evaluation uses greedy decoding on the held-out split under a matched interaction budget. Tokens are per-question means rounded to integers; observation tokens are bounded by the per-round observation cap, so their spread across methods is compressed relative to marker counts. \Cref{tab:resource_accounting} reports the corresponding 7B audit.}
\label{tab:cost_audit_nq}
\footnotesize
\setlength{\tabcolsep}{4pt}
\renewcommand{\arraystretch}{1.06}
\begin{adjustbox}{max width=\linewidth}
\begin{tabular}{lcccc}
\toprule
Method & EM$\uparrow$ & Search/Q & Gen.\ tok/Q & Obs.\ tok/Q \\
\midrule
Base            & 0.142 & 0.000 & 30    & 0     \\
Prompted Search & 0.250 & 2.215 & 120   & 500   \\
\rowcolor{gray!5}
Search-R1       & 0.352 & 3.281 & 149   & 510   \\
\rowcolor{cyan!10}
Search-G1       & 0.453 & 2.094 & \textbf{85} & \textbf{468} \\
\bottomrule
\end{tabular}
\end{adjustbox}
\end{table}

\begin{table}[tbp]
\centering
\caption{Comparison of search and token costs on NQ with Qwen2.5-7B-Instruct. Evaluation uses greedy decoding on the held-out split under a matched interaction budget, as in the 3B audit in~\Cref{tab:cost_audit_nq}. Tokens are per question.}
\label{tab:resource_accounting}
\footnotesize
\setlength{\tabcolsep}{3.6pt}
\renewcommand{\arraystretch}{1.06}
\begin{adjustbox}{max width=\linewidth}
\begin{tabular}{lcccc}
\toprule
Method & EM$\uparrow$ & Search/Q & Gen.\ tok/Q$\downarrow$ & Obs.\ tok/Q$\downarrow$ \\
\midrule
Base            & 0.212 & 0.000 & 28  & 0   \\
Prompted Search & 0.352 & 2.171 & 110 & 490 \\
\rowcolor{gray!5}
Search-R1       & 0.417 & 3.090 & 140 & 500 \\
\rowcolor{cyan!10}
Search-G1       & 0.512 & 1.860 & \textbf{78} & \textbf{440} \\
\bottomrule
\end{tabular}
\end{adjustbox}
\end{table}

\paragraph{Grounding metric definitions.}
TC's searched-trajectory component and D-Inv use a document-level counterfactual intervention on originally correct searched trajectories. The evaluator replaces the gold answer in answer-bearing retrieved documents with an alternative entity or value, then regenerates an answer from the modified documents. Every evaluated correct searched trajectory is assigned exactly one class: \textsc{CF-follow} when its normalized answer changes to track the injected counterfactual entity or value, \textsc{Invariant} when it remains invariant and preserves the original correct answer, and \textsc{Other} otherwise. This evaluation protocol is distinct from the evidence-deletion labels used to fit the evidence-reliance readout: the training-time label perturbs evidence by \emph{deletion} and is used only to fit $D_{\phi^{(m)}}$, while TC/D-Inv perturb evidence by \emph{document-level entity replacement} and are computed post hoc without feeding back into the reward.

\paragraph{Proxy scope and triangulation.}
The deletion intervention retains the reasoning prefix and compares a sampled realized answer with a deterministic counterfactual decode, thereby measuring answer-stage sensitivity conditional on that prefix. \Cref{tab:label_validity} reports the stricter prefix-deletion audit, while \Cref{tab:label_audit} reports paired-decoding controls. Document replacement supplies a non-identical held-out intervention with no shared gradient path, while the independent human evaluation uses neither perturbation nor learned readout. Agreement across these views tests whether the grounding pattern persists across distinct operationalizations of evidence use rather than being inherited from the training label.

For a search-enabled policy with at least one classified correct searched trajectory, let $\mathcal C$ be the union of originally correct direct trajectories and originally correct searched trajectories classified by the intervention. \textbf{TC} is a graded score of correct-direct retention and counterfactual response
\[
\mathrm{TC}=\frac{1}{|\mathcal C|}\sum_{\tau\in\mathcal C}s(\tau),
\qquad
s(\tau)\in\{1.0,0.6,0.3\}.
\]
Correct direct trajectories receive score $1.0$ as the selective-answering component; classified correct searched trajectories receive $1.0$ for \textsc{CF-follow}, $0.6$ for \textsc{Other}, and $0.3$ for \textsc{Invariant}. Thus TC is a joint search-policy measure of correct direct retention and counterfactual response. Because its absolute level depends on fixed heuristic weights and the direct-answer convention, we interpret it jointly with D-Inv, a weight-free searched-only proportion, and with the S-TC control reported in~\Cref{tab:efficiency_controlled}, which removes direct trajectories from the denominator. Search-G1 retains the leading value under these companion endpoints, so its qualitative advantage is not driven solely by TC's direct-answer convention. TC and D-Inv are undefined for the search-disabled Base reference because no retrieved context is available for the intervention protocol. \textbf{D-Inv} is counterfactual-document invariance among classified originally correct searched trajectories,
\[
\mathrm{D\text{-}Inv}
=
\frac{N_{\mathrm{Invariant}}}
{N_{\mathrm{CF\text{-}follow}}+N_{\mathrm{Invariant}}+N_{\mathrm{Other}}},
\]
so lower is better. D-Inv contains no direct-answer term and therefore avoids TC's mechanical direct-answer contribution, although it remains conditional on each policy's correct searched trajectories. \Cref{tab:main} shows the NQ comparison, and \Cref{tab:grounding_2wiki} shows the same D-Inv/TC ordering on 2Wiki. We read D-Inv as the primary evidence-dependence endpoint and TC as a summary of correct-direct retention and counterfactual response; S-TC provides the complementary searched-only control for TC's composition effect.

Document replacement is an operational diagnostic of counterfactual evidence dependence, not a necessary-and-sufficient test of grounding. Under answer-bearing-fact replacement, invariance is evidence of weak dependence on the manipulated fact, whereas responsiveness may reflect evidential reasoning, surface context following, or perturbation-induced failure. Moving an invariant answer to \textsc{Other} also lowers D-Inv. We therefore interpret D-Inv jointly with the measured surface controls in~\Cref{tab:shortcut_controls}, the independent human support evaluation in~\Cref{tab:human_grounding}, and the common-question analysis in~\Cref{app:decomposition}. Unless noted, \Cref{tab:main} reports TC and D-Inv on NQ.

\paragraph{Readout-audit provenance.}
The NQ--3B representation result in~\Cref{fig:estimator_fidelity} uses layer 27 on $3{,}663$ correct trajectories ($2{,}015$ sensitive; $1{,}648$ invariant), separate from the scalar/logit-control pool. \Cref{fig:drift} uses Qwen2.5-3B/7B-Instruct on 2WikiMultiHopQA; the 3B series covers checkpoints $0,25,50,75,100,150$.

\begin{figure}[tbp]
\centering
\includegraphics[width=0.92\textwidth]{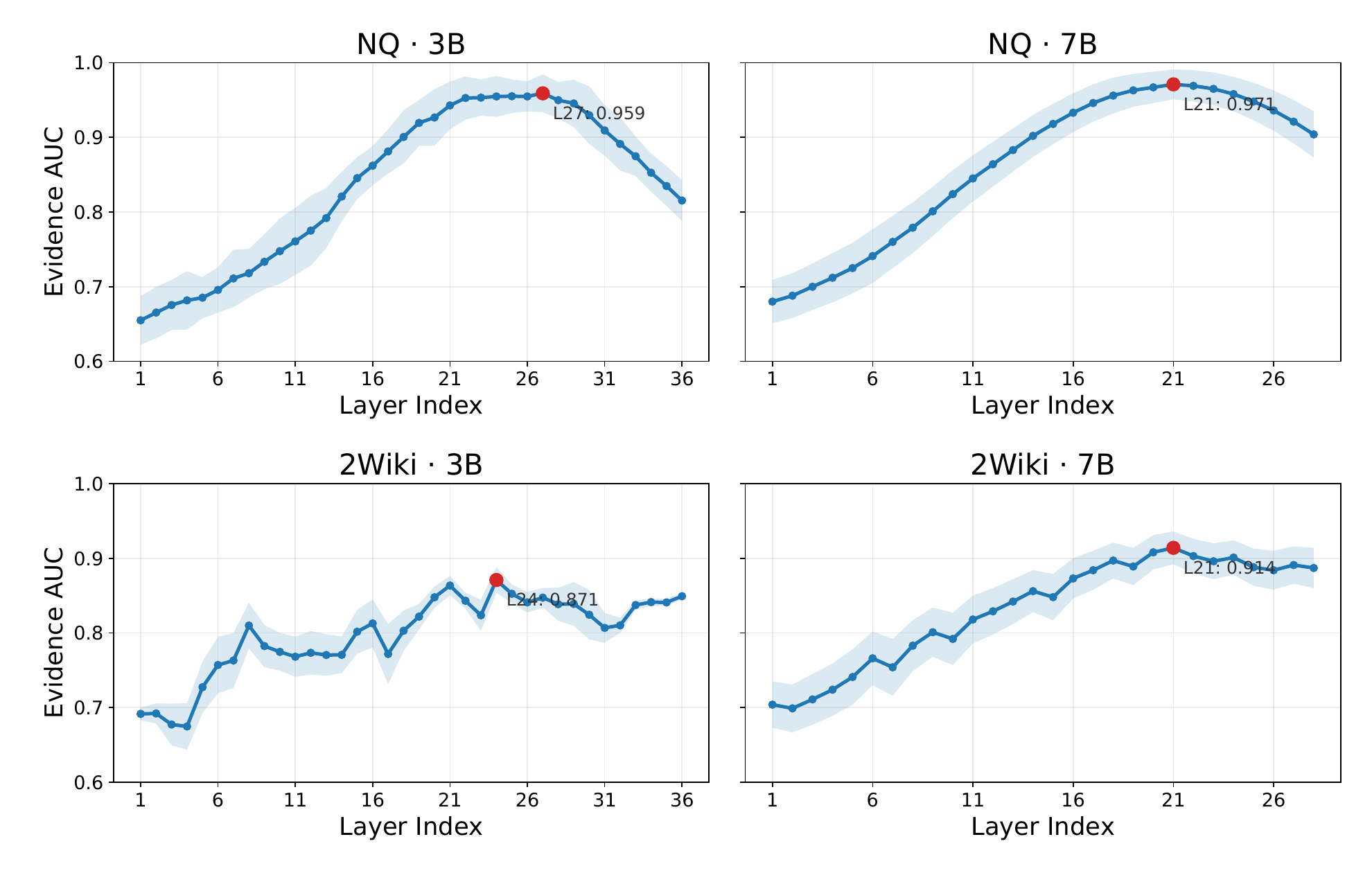}
\caption{Layer-wise linear separability of evidence-deletion sensitivity on NQ and 2WikiMultiHopQA using Qwen2.5-3B and Qwen2.5-7B. Each point reports commit-token readout AUC among $\mathrm{EM}=1$ rollouts.}
\label{fig:layer_localization}
\end{figure}

\paragraph{Layer-wise linear separability.}
We further examine where the deletion-sensitivity label is most linearly separable. For each layer, we train the same linear readout to distinguish deletion-sensitive from deletion-invariant correct trajectories and compare separability across depth. As shown in~\Cref{fig:layer_localization}, the best layer varies across settings---layer 27/21 on NQ and 24/21 on 2Wiki for 3B/7B---supporting layer selection by model and refit rather than one universal depth.

\paragraph{Counterfactual label reliability.}
As defined in~\Cref{eq:evidence_reliance_def}, Search-G1 operationalizes evidence reliance by whether the realized answer changes under evidence ablation. The 3B and 7B NQ/HotpotQA rows compare the default deletion label against a stricter intervention that also removes intermediate reasoning text. Agreement is approximately 0.95 with $\kappa\approx0.88$, showing that reasoning-prefix leakage is limited in the audited settings.

\begin{table}[tbp]
\centering
\caption{Agreement between default and strict evidence-deletion labels on NQ and HotpotQA at 3B and 7B.}
\label{tab:label_validity}
\footnotesize
\setlength{\tabcolsep}{4.8pt}
\renewcommand{\arraystretch}{1.06}
\begin{adjustbox}{max width=\linewidth}
\begin{tabular}{lcccc}
\toprule
Dataset & Agree.$\uparrow$ & $\kappa\uparrow$ & \makecell{Reliance\\rate} & \makecell{Strict\\reliance rate} \\
\midrule
\multicolumn{5}{c}{Qwen2.5-3B-Instruct} \\
\midrule
NQ & 0.953 & 0.881 & 0.724 & 0.747 \\
\rowcolor{gray!5}
HotpotQA & 0.952 & 0.879 & 0.712 & 0.741 \\
\midrule
\multicolumn{5}{c}{Qwen2.5-7B-Instruct} \\
\midrule
NQ
& 0.956
& 0.897
& 0.681
& 0.704 \\
\rowcolor{gray!5}
HotpotQA
& 0.954
& 0.894
& 0.668
& 0.696 \\
\bottomrule
\end{tabular}
\end{adjustbox}
\end{table}

\paragraph{Metric definitions for the closed-book analyses.}
We define the HotpotQA metrics in~\Cref{tab:search_efficiency_cb} and the 2Wiki metrics in~\Cref{tab:cbwrong_retrieval}. The former defines CB-correct separately per policy from its deterministic retrieval-disabled decode; the latter fixes Base-model CB-wrong within each scale. \textbf{CB-EM}/\textbf{CBw-EM} are exact match on these conditional sets, while \textbf{CB-S/Q}/\textbf{All-S/Q} count response-side search markers per question on the CB-correct/full audited slice. \textbf{Over-S} is the valid-trajectory fraction with multiple markers; \textbf{Any@3} is the evaluated CB-wrong fraction with at least one correct sampled trajectory. These metrics condition on valid parsed answers, so counts may differ across methods. \textbf{Valid-S/Q} names this conditional 2Wiki statistic; \Cref{tab:grounding_2wiki,tab:efficiency_controlled} use all-question Search/Q. None uses evidence perturbation.

\begin{table}[tbp]
\centering
\caption{Known-answer preservation and search behavior on a 1,000-question HotpotQA slice. CB-correct is defined separately for each policy from its retrieval-disabled decode.}
\label{tab:search_efficiency_cb}
\footnotesize
\setlength{\tabcolsep}{5.5pt}
\renewcommand{\arraystretch}{1.06}
\begin{adjustbox}{max width=\linewidth}
\begin{tabular}{lcccc}
\toprule
Method & CB-EM$\uparrow$ & CB-S/Q$\downarrow$ & All-S/Q$\downarrow$ & Over-S$\downarrow$ \\
\midrule
\multicolumn{5}{c}{Qwen2.5-3B-Instruct} \\
\midrule
Prompted Search & 0.380 & 1.199 & 1.215 & 0.225 \\
\rowcolor{gray!5}
Search-R1 & 0.541 & 1.535 & 1.629 & 0.489 \\
\rowcolor{cyan!10}
Search-G1 & \textbf{0.573} & \textbf{1.056} & \textbf{1.049} & \textbf{0.040} \\
\midrule
\multicolumn{5}{c}{Qwen2.5-7B-Instruct} \\
\midrule
Prompted Search & 0.431 & 1.085 & 1.102 & 0.194 \\
\rowcolor{gray!5}
Search-R1 & 0.592 & 1.411 & 1.503 & 0.438 \\
\rowcolor{cyan!10}
Search-G1 & \textbf{0.641} & \textbf{0.914} & \textbf{0.905} & \textbf{0.028} \\
\bottomrule
\end{tabular}
\end{adjustbox}
\end{table}

\begin{table}[tbp]
\centering
\caption{Retrieval performance on fixed Base-model closed-book-wrong subsets of 2WikiMultiHopQA. Metrics condition on valid trajectories; methods share question IDs within each scale.}
\label{tab:cbwrong_retrieval}
\footnotesize
\setlength{\tabcolsep}{5.5pt}
\renewcommand{\arraystretch}{1.06}
\begin{adjustbox}{max width=\linewidth}
\begin{tabular}{lcccc}
\toprule
Method & CBw-EM$\uparrow$ & Any@3$\uparrow$ & Valid-S/Q$\downarrow$ & Over-S$\downarrow$ \\
\midrule
\multicolumn{5}{c}{Qwen2.5-3B-Instruct} \\
\midrule
Prompted Search & 0.145 & 0.236 & 1.129 & 0.315 \\
\rowcolor{gray!5}
Search-R1 & 0.254 & 0.333 & 1.972 & 0.809 \\
\rowcolor{cyan!10}
Search-G1 & \textbf{0.327} & \textbf{0.435} & 1.649 & 0.605 \\
\midrule
\multicolumn{5}{c}{Qwen2.5-7B-Instruct} \\
\midrule
Prompted Search & 0.162 & 0.252 & 1.050 & 0.284 \\
\rowcolor{gray!5}
Search-R1 & 0.278 & 0.362 & 1.840 & 0.748 \\
\rowcolor{cyan!10}
Search-G1 & \textbf{0.356} & \textbf{0.476} & 1.520 & 0.548 \\
\bottomrule
\end{tabular}
\end{adjustbox}
\end{table}

\paragraph{Search-action frequency and known-answer preservation.}
We test whether retrieval training preserves answers within each policy's estimated knowledge boundary. \Cref{tab:search_efficiency_cb} shows that, on the audited HotpotQA slice, Search-G1 has the highest CB-EM and the lowest CB-S/Q and All-S/Q among the compared methods. Relative to Search-R1, CB-EM is 0.573 versus 0.541 at 3B and 0.641 versus 0.592 at 7B; All-S/Q is 1.049 versus 1.629 and 0.905 versus 1.503. Because each method defines its own CB-correct set, the following analysis complements this view with a fixed Base-model CB-wrong set on 2Wiki.

\paragraph{Closed-book-wrong question performance.}
We next evaluate the CB-wrong subset, on which the base model fails without retrieval. \Cref{tab:cbwrong_retrieval} shows that, on 2WikiMultiHopQA, Search-G1 reaches mean valid-conditioned EM $0.327/0.356$ versus $0.254/0.278$ for Search-R1 at 3B/7B, with Valid-S/Q $1.649/1.520$ versus $1.972/1.840$. Thus, in this subset, lower mean search-marker frequency co-occurs with higher mean valid-conditioned EM.

\paragraph{Cross-dataset transfer.}
\Cref{tab:cross_dataset_transfer} maps each source-trained Search-G1 policy to all four held-out targets. Across scales, $23/24$ off-diagonal means exceed Prompted Search. For NQ-trained Search-G1 versus NQ-trained Search-R1, paired EM differences [95\% CI] are $+0.101\,[+0.072,+0.130]$, $+0.008\,[-0.012,+0.028]$, $+0.006\,[-0.015,+0.027]$, and $+0.001\,[-0.012,+0.014]$ on NQ, HotpotQA, 2Wiki, and MuSiQue, respectively. The largest and clearest gain is in-domain; the smaller out-of-domain differences have intervals spanning zero.

\begin{table}[tbp]
\centering
\caption{In-domain and cross-dataset transfer of Search-G1 across four QA benchmarks at 3B and 7B. Each cell reports EM / Search/Q; diagonal cells are in-domain and off-diagonal cells are cross-dataset. Grounding comparisons focus on NQ and 2Wiki in Tables~\ref{tab:main} and~\ref{tab:grounding_2wiki}.}
\label{tab:cross_dataset_transfer}
\footnotesize
\setlength{\tabcolsep}{4.0pt}
\renewcommand{\arraystretch}{1.08}
\begin{adjustbox}{max width=\linewidth}
\begin{tabular}{lcccc}
\toprule
\multirow{2}{*}{Training Dataset}
&
\multicolumn{4}{c}{Target Test Dataset}
\\
\cmidrule(lr){2-5}
&
NQ & HotpotQA & 2Wiki & MuSiQue
\\
\midrule
\multicolumn{5}{c}{Qwen2.5-3B-Instruct} \\
\midrule
Prompted Search
& 0.250 / 2.215
& 0.199 / 2.531
& 0.238 / 2.951
& 0.057 / 2.141 \\

NQ
& \textbf{0.453 / 2.094}
& 0.248 / 2.480
& 0.256 / 2.551
&  0.091 / 2.406 \\
\rowcolor{gray!5}
HotpotQA
& 0.359 / 2.205
& \textbf{0.306 / 2.549}
& 0.270 / 2.828
& 0.099 / 2.881 \\
2Wiki
& 0.377 / 2.486
& 0.316 / 2.439
& \textbf{0.368 / 2.812}
& 0.116 / 3.275 \\
\rowcolor{gray!5}
MuSiQue
& 0.297 / 3.640
& 0.241 / 3.389
& 0.279 / 3.227
& \textbf{0.143 / 3.443} \\
\midrule
\multicolumn{5}{c}{Qwen2.5-7B-Instruct} \\
\midrule
Prompted Search
& 0.352 / 2.171
& 0.292 / 2.480
& 0.274 / 2.892
& 0.074 / 2.098 \\
NQ
& \textbf{0.512 / 1.860}
& 0.365 / 2.202
& 0.330 / 2.265
& 0.162 / 2.137 \\
\rowcolor{gray!5}
HotpotQA
& 0.406 / 1.958
& \textbf{0.450 / 2.263}
& 0.349 / 2.511
& 0.176 / 2.558 \\
2Wiki
& 0.426 / 2.208
& 0.465 / 2.166
& \textbf{0.475 / 2.495}
& 0.206 / 2.908 \\
\rowcolor{gray!5}
MuSiQue
& 0.336 / 3.232
& 0.354 / 3.009
& 0.360 / 2.866
& \textbf{0.254 / 3.057} \\
\bottomrule
\end{tabular}
\end{adjustbox}
\end{table}

\paragraph{Independent human grounding evaluation.}
\Cref{sec:analysis} summarizes this evaluation; here we give the per-method table, the annotation protocol, and the agreement statistics. Each pool contains $180$ correct searched instances drawn independently per method, so answer correctness is fixed by construction and annotators judge only whether the retrieved evidence supports the answer.

\begin{table}[tbp]
\centering
\caption{Human evaluation of answer support on independent 180-instance correct-searched pools per method with Qwen2.5-3B. S@C is mean normalized support; Full and Uns./CBU are the fully supported and unsupported fractions. This evaluation uses no evidence perturbation, re-decoding, or learned readout.}
\label{tab:human_grounding}
\footnotesize
\setlength{\tabcolsep}{4pt}
\renewcommand{\arraystretch}{1.08}
\begin{adjustbox}{max width=\linewidth}
\begin{tabular}{lccc}
\toprule
Method & S@C\,($\uparrow$) & Full\,($\uparrow$) & Uns./CBU\,($\downarrow$) \\
\midrule
Prompted Search      & 0.594 & 0.400 & 0.211 \\
\rowcolor{gray!5}
Search-R1            & 0.650 & 0.472 & 0.172 \\
GPT-5 Judge  & 0.797 & 0.678 & 0.083 \\
\rowcolor{gray!5}
Search-E1            & 0.719 & 0.561 & 0.122 \\
\makecell[l]{w/o Evidence-Reliance\\Reward}  & 0.633 & 0.450 & 0.183 \\
\rowcolor{cyan!10}
\textbf{Search-G1 (ours)} & \textbf{0.842} & \textbf{0.733} & \textbf{0.050} \\
\bottomrule
\end{tabular}
\end{adjustbox}
\end{table}

\textit{Annotation.}
Three annotators independently label each instance on a $3$-point support scale:
(i)~\textbf{Fully supported} ($2$): the evidence explicitly states, or unambiguously entails through a short and complete reasoning chain, every fact required for the answer;
(ii)~\textbf{Partially supported} ($1$): the evidence supports only part of the answer, covers only some hops, or is relevant but insufficient to entail the answer;
(iii)~\textbf{Unsupported} ($0$): the evidence is irrelevant, contradicts the answer, or does not contain enough information to support the answer.
The final label is the majority vote; all-disagreement cases are adjudicated by a fourth senior annotator.

\textit{Target measures.}
We report \textbf{Support@Correct} (S@C), the mean majority-vote support score normalized to $[0,1]$; \textbf{Fully Supported} (Full), the fraction of score-$2$ instances; and \textbf{Unsupported} (Uns.), the fraction of score-$0$ instances. Since all evaluated trajectories are exact-match correct, Uns. is also the \emph{Correct-but-Unsupported} (CBU) rate: the fraction of correct answers not supported by the retrieved evidence shown to annotators. This evaluation uses no evidence-deletion labels, counterfactual document replacement, re-decoding, or Search-G1 reward readouts.

\textit{Agreement.}
The annotation reaches substantial pre-adjudication agreement (Fleiss' $\kappa=0.79$, ordinal Krippendorff's $\alpha=0.81$ on Qwen2.5-3B-Instruct).

\Cref{tab:human_grounding} compares six methods on distinct correct-searched pools. The unconditional common-question decomposition in~\Cref{app:decomposition} provides a complementary view on common question IDs.

\subsubsection{Sensitivity and controlled comparisons}
\label{app:ablation}
\paragraph{Shared anchors and variant definitions.}
The component, coefficient, refresh, and unified-accounting studies below reuse the same four trained policies as comparison anchors rather than reporting additional repetitions. The $\eta=0$ row is the \emph{w/o search cost} condition, $K=\infty$ is the no-refresh evaluator condition, and \emph{w/o searched gate} is the main no-gate condition. In the latter, $n_i$ is replaced by $1$ only in the correct-searched reliance term; $b_i$ and all other branches remain unchanged. The four-round Search-G1 entry is the default interaction-budget anchor.

\paragraph{Validation-based hyperparameter selection.}
\Cref{tab:hyperparam_selection} summarizes Search-G1 coefficient and refresh sweeps together with the baseline-by-baseline tuning ledger. The same validation rule is used for every method: retain configurations within 0.02 EM of the best validation EM, maximize TC, then minimize Search/Q; test data are never consulted.

\begin{table}[tbp]
\centering
\caption{Hyperparameter candidates and validation-selected settings for Search-G1 and baseline methods. Candidate counts and per-method selections use the shared validation rule; Search-G1's default matches the reported coefficient/update sweeps. ``Scale'' is relative to the official auxiliary-reward coefficient.}
\label{tab:hyperparam_selection}
\footnotesize
\setlength{\tabcolsep}{4.2pt}
\renewcommand{\arraystretch}{1.06}
\begin{adjustbox}{max width=\linewidth}
\begin{tabular}{llcl}
\toprule
Method & Tuned quantities & Configs per scale & Selected setting (3B / 7B) \\
\midrule
Search-G1 & $\lambda_g,\eta,K,M_{\mathrm{cal}}$ & 12 & $(0.6,0.02,25,300)$ / same \\
Search-R1 & direct bonus; KL coefficient & 9 & $(0.5,10^{-3})$ / same \\
GiGPO & state-credit scale; anchor-group cap & 9 & $(1.0,5)$ / same \\
AgentPRM & process-reward scale; KL coefficient & 9 & $(1.0,10^{-3})$ / same \\
GPT-5 Judge & support-reward scale; KL coefficient & 9 & $(0.5,10^{-3})$ / same \\
ARPO & entropy scale; allocation temperature & 9 & $(1.0,1.0)$ / same \\
IGPO & information-gain scale; normalization & 9 & $(1.0,\text{group})$ / same \\
Search-E1 & distillation mixture; temperature & 9 & $(0.25,1.0)$ / same \\
SAAS & boundary threshold; allocation scale & 9 & $(0.5,1.0)$ / same \\
$\beta$-GRPO & confidence threshold; reward scale & 9 & $(0.5,1.0)$ / same \\
KbPO & certainty weight; quality weight & 9 & $(0.5,0.5)$ / same \\
\bottomrule
\end{tabular}
\end{adjustbox}
\end{table}

\paragraph{Reward coefficient sensitivity.}
We study sensitivity to the evidence-reliance and search-cost coefficients on 2WikiMultiHopQA. \Cref{tab:coef_sensitivity} varies $\lambda_g$ and $\eta$ one at a time while keeping the other training settings fixed. Among tested values, the default $(0.6,0.02)$ gives the highest TC and lowest Search/Q at both scales; $\lambda_g=0.8$ gives higher EM but worse TC and Search/Q, and increasing $\eta$ to $0.05$ increases Search/Q in these runs. We fix $\alpha_{\mathrm{cb}}=0.5$, matching the maximum correct-direct bonus of the Search-R1 reference. \Cref{eq:direct_search_order} characterizes the local pre-normalization incentive; after retraining, policy-level Search/Q need not vary monotonically with an individual coefficient.

\begin{table}[tbp]
\centering
\caption{Sensitivity to evidence-reliance and search-cost coefficients on 2WikiMultiHopQA.}
\label{tab:coef_sensitivity}
\footnotesize
\setlength{\tabcolsep}{5pt}
\renewcommand{\arraystretch}{1.06}
\begin{tabular}{ccccc}
\toprule
$\lambda_g$ & $\eta$ & EM$\uparrow$ & TC & Search/Q$\downarrow$ \\
\midrule
\multicolumn{5}{c}{Qwen2.5-3B-Instruct} \\
\midrule
0.4 & 0.02 & 0.347& 0.854 & 2.842 \\
\rowcolor{cyan!10}
0.6 & 0.02 & 0.368 & 0.972 & 2.812 \\
0.8 & 0.02 & 0.381& 0.911 & 3.383
\\
\rowcolor{gray!5}
0.6 & 0.00 & 0.338 & 0.920 & 3.576 \\
0.6 & 0.05 & 0.334 & 0.891 & 3.568
\\
\midrule
\multicolumn{5}{c}{Qwen2.5-7B-Instruct} \\
\midrule
0.4 & 0.02
& 0.457
& 0.878
& 2.521 \\
\rowcolor{cyan!10}
0.6 & 0.02
& 0.475
& 0.978
& 2.495 \\
0.8 & 0.02
& 0.486
& 0.926
& 2.980 \\
\rowcolor{gray!5}
0.6 & 0.00
& 0.449
& 0.934
& 3.144 \\
0.6 & 0.05
& 0.446
& 0.909
& 3.138 \\
\bottomrule
\end{tabular}
\end{table}

\paragraph{Estimator update sensitivity.}
We study how often the Search-G1 estimators should be refit and how many refitting questions are needed at each update. As the policy changes, the targets and trajectory-conditioned state distributions move away from the initial calibration data. \Cref{tab:update_sensitivity} evaluates different co-evolution schedules at both model scales on 2WikiMultiHopQA, where $K$ is the refitting interval and $M_{\mathrm{cal}}$ is the number of refitting questions. Among tested schedules, $K=25,M_{\mathrm{cal}}=300$ gives the highest EM and TC; $K=50$ lowers Search/Q with slightly lower outcomes, while $K=10$ or $M_{\mathrm{cal}}=100$ gives a weaker joint profile.

\begin{table}[tbp]
\centering
\caption{Sensitivity to readout refresh frequency and calibration-set size on 2WikiMultiHopQA.}
\label{tab:update_sensitivity}
\footnotesize
\setlength{\tabcolsep}{3pt}
\renewcommand{\arraystretch}{1.06}
\begin{adjustbox}{max width=\linewidth}
\begin{tabular}{lccccc}
\toprule
\makecell[l]{Update\\schedule} & $K$ & $M_{\mathrm{cal}}$ & EM$\uparrow$ & TC & \makecell{Search/Q\\$\downarrow$} \\
\midrule

\multicolumn{6}{c}{Qwen2.5-3B-Instruct} \\
\midrule
No-refresh evaluator
& $\infty$ & 0
& 0.357 & 0.906 & 3.225 \\

\makecell[l]{Nominal 50-update\\cadence}
& 50 & 300
& 0.360 & 0.937 & 2.656 \\

\rowcolor{cyan!10}
\makecell[l]{Nominal 25-update\\cadence}
& 25 & 300
& 0.368 & 0.972 & 2.812 \\

\makecell[l]{Nominal 10-update\\cadence}
& 10 & 300
& 0.333 & 0.895 & 3.589 \\

\rowcolor{gray!5}
\makecell[l]{Nominal 25-update\\cadence}
& 25 & 100
& 0.339 & 0.912 & 3.839 \\

\midrule
\multicolumn{6}{c}{Qwen2.5-7B-Instruct} \\
\midrule
No-refresh evaluator
& $\infty$ & 0
& 0.466
& 0.922
& 2.846 \\

\makecell[l]{Nominal 50-update\\cadence}
& 50 & 300
& 0.468
& 0.948
& 2.362 \\

\rowcolor{cyan!10}
\makecell[l]{Nominal 25-update\\cadence}
& 25 & 300
& 0.475
& 0.978
& 2.495 \\

\makecell[l]{Nominal 10-update\\cadence}
& 10 & 300
& 0.445
& 0.913
& 3.155 \\

\rowcolor{gray!5}
\makecell[l]{Nominal 25-update\\cadence}
& 25 & 100
& 0.450
& 0.927
& 3.368 \\

\bottomrule
\end{tabular}
\end{adjustbox}
\end{table}

\paragraph{Common-subset comparison under unified accounting.}
We score all methods on one fixed 1{,}024-question 2WikiMultiHopQA subset under unified accounting: identical question IDs, invalid trajectories charged EM${=}0$, and Search/Q and per-marker EM computed over all questions. To avoid overloading the main TC definition, we denote the searched-trajectory-only counterfactual-response score by \textbf{S-TC}; it averages the same counterfactual scores over correct searched trajectories only. Search-G1 attains the highest all-question EM, S-TC, and EM per search marker. This table aligns question IDs and marker accounting. \Cref{tab:cost_audit_nq} gives the matched 3B NQ token audit, and \Cref{tab:resource_accounting} gives the 7B counterpart.

\begin{table}[tbp]
\centering
\caption{Comparison under unified accounting on a fixed 1,024-question 2WikiMultiHopQA subset with Qwen2.5-3B. S-TC is the searched-trajectory-only counterfactual-response score and is distinct from the union-of-correct TC. Invalid trajectories are charged EM$=0$; Search/Q is averaged over all 1,024 IDs.}
\label{tab:efficiency_controlled}
\footnotesize
\setlength{\tabcolsep}{5.0pt}
\renewcommand{\arraystretch}{1.08}
\begin{adjustbox}{max width=\linewidth}
\begin{tabular}{lcccc}
\toprule
Method & EM$\uparrow$ & S-TC$\uparrow$ & Search/Q$\downarrow$ & EM per marker$\uparrow$ \\
\midrule
Prompted Search & 0.231 & 0.608 & 1.602 & 0.144 \\
\rowcolor{gray!5}
Search-R1       & 0.285 & 0.534 & 1.942 & 0.147 \\
\rowcolor{cyan!10}
Search-G1       & \textbf{0.355} & \textbf{0.655} & \textbf{1.580} & \textbf{0.225} \\
\bottomrule
\end{tabular}
\end{adjustbox}
\end{table}

\subsubsection{Case study}
\label{app:case}

We provide selected qualitative question-level summaries from Qwen2.5-3B-Instruct on 2WikiMultiHopQA. Each case summarizes representative outputs for a question on which both Prompted Search and the Search-R1 baseline fail while Search-G1 answers correctly. The cases illustrate method-level differences in task accuracy, evidence alignment, and search behavior under this selection rule.

\begin{figure}[tbp]
\centering
\includegraphics[width=\textwidth]{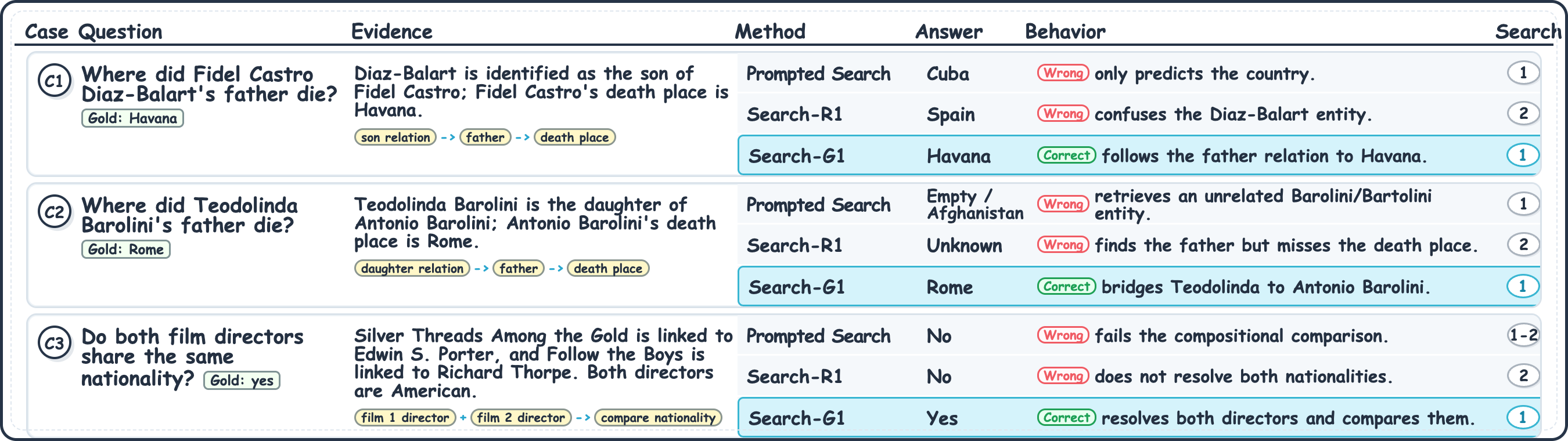}
\caption{Qualitative comparison of selected search trajectories and evidence use on 2WikiMultiHopQA with Qwen2.5-3B-Instruct. The Evidence column gives a reference chain for interpreting each question; method rows summarize representative observed outputs rather than implying identical retrieved contexts.}
\label{fig:casestudy}
\end{figure}

\subsubsection{Additional validity and robustness checks}
\label{app:audits}

\paragraph{Evidence-reliance label audit.}
\label{app:label_audit}
The evidence-reliance label in~\Cref{eq:evidence_reliance_def} compares a sampled trajectory answer with a deterministic evidence-ablated decode, so it can mix evidence sensitivity with decoding mismatch. We therefore re-derive the label under (i) \emph{paired-greedy} decoding and (ii) \emph{common-random-number (CRN) paired sampling}, which couples the sampling variates to reduce independent decoding noise. On Qwen2.5-3B-Instruct/NQ at checkpoint 50, the original labels agree with these alternatives on 93\%/96\% of trajectories ($\kappa=0.85/0.90$). As reported in~\Cref{tab:label_audit}, the positive-label rate changes by $-0.02/-0.01$, while 6\%/4\% of group-relative advantage signs flip. Thus decoding mismatch is not the dominant source of the label in this audit, although its optimization effect is nonzero.

\begin{table}[tbp]
\centering
\caption{Agreement of evidence-reliance labels under paired decoding controls on NQ with Qwen2.5-3B-Instruct. Controls use paired-greedy decoding and common-random-number paired sampling.}
\label{tab:label_audit}
\footnotesize
\begin{adjustbox}{max width=\linewidth}
\begin{tabular}{lcccc}
\toprule
Construction & Agree.$\uparrow$ & $\kappa\uparrow$ & Pos.-rate $\Delta$ & Adv.-sign flip$\downarrow$ \\
\midrule
Paired-greedy      & 0.93 & 0.85 & $-0.02$ & 0.06 \\
\rowcolor{gray!5}
CRN paired sampling & 0.96 & 0.90 & $-0.01$ & 0.04 \\
\bottomrule
\end{tabular}
\end{adjustbox}
\end{table}

\paragraph{Shortcut-control audit.}
\Cref{tab:shortcut_controls} reports supplementary readout diagnostics using surface features, shuffled labels, and matching on search count and response length.

\begin{table}[tbp]
\centering
\caption{Evidence-reliance prediction under surface-feature and label-shuffling controls on NQ and 2WikiMultiHopQA with Qwen2.5-3B-Instruct.}
\label{tab:shortcut_controls}
\footnotesize
\setlength{\tabcolsep}{5.2pt}
\renewcommand{\arraystretch}{1.06}
\begin{adjustbox}{max width=\linewidth}
\begin{tabular}{lcc}
\toprule
Predictor/control & NQ AUC & 2Wiki AUC \\
\midrule
Search count only & 0.548 & 0.562 \\
Response length only & 0.557 & 0.574 \\
Retrieved-token count only & 0.536 & 0.551 \\
Answer--evidence lexical overlap & 0.589 & 0.603 \\
\rowcolor{gray!5}
Label-shuffled representation & 0.501 & 0.498 \\
\rowcolor{cyan!10}
Full answer-commit representation & 0.959 & 0.871 \\
Matched on search count and length & 0.934 & 0.842 \\
\bottomrule
\end{tabular}
\end{adjustbox}
\end{table}

\paragraph{Retrieval-necessity label audit.}
\label{app:necessity_audit}
The necessity target $z_i$ in~\Cref{eq:necessity_target} comes from a deterministic closed-book decode scored by normalized answer inclusion, which is a more permissive criterion than the task exact-match verifier. On the same full held-out Base-model decodes used in~\Cref{tab:boundary_composition}, exact-match correctness is a strict subset of inclusion sufficiency, so the two labels agree on $93.10\%$ of NQ and $96.42\%$ of 2Wiki examples and disagree on $249/3{,}610$ and $450/12{,}576$ respectively.

Relative to EM labeling, inclusion assigns closed-book sufficiency to an additional $6.90\%$ of NQ questions. These disagreements shift calibration targets toward sufficiency. When reflected in higher fitted $b_i$, the broader sufficiency labels favor direct-answer credit under~\Cref{eq:direct_search_order}. \Cref{tab:search_efficiency_cb} characterizes policy-relative CB-correct slices, and \Cref{tab:cbwrong_retrieval} provides the fixed Base-CB-wrong view. We use inclusion because it is robust to free-form closed-book decodes and report its agreement and disagreement counts with EM for transparency.

\paragraph{Necessity-score distribution across refits.}
We recompute $n_i=1-b_i$ by applying each refitted boundary readout to its matched hidden-state dump on the same 300-question 2Wiki calibration slice. \Cref{tab:necessity_score_drift} shows a strongly bimodal distribution across refits: the 10th percentile stays near zero, the median above $0.93$, and the 90th percentile near one. The mean stays within $0.605$--$0.622$ and the predicted-necessary fraction within $59.67$--$61.67\%$. The aggregate boundary composition is therefore stable across these refits, while each checkpoint uses a newly calibrated readout.

\begin{table}[tbp]
\centering
\caption{Retrieval-necessity score distributions across refits on a fixed 300-question 2WikiMultiHopQA calibration slice with Qwen2.5-3B-Instruct. Hard S/N counts closed-book-sufficient/necessary labels; Pred. N is the number of readout scores with $n_i\geq0.5$.}
\label{tab:necessity_score_drift}
\footnotesize
\setlength{\tabcolsep}{3.2pt}
\renewcommand{\arraystretch}{1.06}
\begin{adjustbox}{max width=\linewidth}
\begin{tabular}{lcccccc}
\toprule
Checkpoint & Hard S/N & Mean & SD & q10 / median / q90 & Pred. N ($\%$) & AUC \\
\midrule
25 & 115/185 & 0.617 & 0.454 & 0.011 / 0.969 / 1.000 & 185 (61.67\%) & 0.866 \\
\rowcolor{gray!5}
50 & 118/182 & 0.622 & 0.430 & 0.026 / 0.932 / 1.000 & 183 (61.00\%) & 0.888 \\
75 & 122/178 & 0.605 & 0.448 & 0.013 / 0.948 / 1.000 & 179 (59.67\%) & 0.881 \\
\bottomrule
\end{tabular}
\end{adjustbox}
\end{table}

\paragraph{Common-question unconditional behavior decomposition.}
\label{app:decomposition}
TC and D-Inv condition on method-specific correct trajectories, so an unconditional common-question analysis is needed to expose selection effects. \Cref{tab:decomposition} evaluates all five methods on the same fixed set of $600$ NQ question IDs, using the full/partial/unsupported support categories defined in the preceding annotation protocol. It keeps partial support separate and reports unconditional grounded accuracy
$\mathrm{UGA}=P(\text{correct direct})+P(\text{correct search, full})+0.5P(\text{correct search, partial})$.

\begin{table}[tbp]
\centering
\caption{Unconditional decomposition of answer correctness and support on 600 shared NQ questions. All five methods use the same sample. Search-full, search-partial, and search-unsupported use the support categories defined in the preceding annotation protocol. The five buckets sum to one; UGA is reported with a question-bootstrap 95\% CI.}
\label{tab:decomposition}
\footnotesize
\begin{adjustbox}{max width=\linewidth}
\begin{tabular}{lcccccc}
\toprule
Method & Corr.-direct & Search-full & Search-partial & Search-unsupp.$\downarrow$ & Incorrect & UGA [95\% CI]$\uparrow$ \\
\midrule
Prompted Search & 0.10 & 0.09 & 0.05 & 0.05 & 0.71 & $0.215\,[0.184,0.248]$ \\
\rowcolor{gray!5}
Search-R1 & 0.09 & 0.15 & 0.07 & 0.06 & 0.63 & $0.275\,[0.240,0.312]$ \\
GPT-5 Judge & 0.11 & 0.21 & 0.07 & 0.03 & 0.58 & $0.355\,[0.317,0.394]$ \\
\rowcolor{gray!5}
Search-E1 & 0.11 & 0.19 & 0.07 & 0.04 & 0.59 & $0.335\,[0.298,0.373]$ \\
\rowcolor{cyan!10}
Search-G1 & 0.12 & 0.20 & 0.08 & 0.03 & 0.57 & $0.360\,[0.322,0.399]$ \\
\bottomrule
\end{tabular}
\end{adjustbox}
\end{table}

On this shared question set, Search-G1 reaches unconditional grounded accuracy $0.360$, compared with $0.355$ for GPT-5 Judge, with the same rounded unsupported rate. This provides an independent common-denominator check that does not use Search-G1's learned reward readouts.

\paragraph{Credit--support association audit.}
Within each original $G=5$ group, we permute $d_{i,j}$ only among correct searched trajectories with identical $c_{i,j}$, keeping other reward terms fixed. Cost-only removes only the correct-searched reliance bonus. Pearson correlations with human support use correct searched trajectories and advantages from complete original groups, with within-group centering. \Cref{tab:credit_control} reports these auxiliary associations and task outcomes.

\begin{table}[tbp]
\centering
\caption{Association between learning credit and human-assessed support on 2WikiMultiHopQA with Qwen2.5-3B-Instruct. Support and correlations include 95\% CIs.}
\label{tab:credit_control}
\footnotesize
\setlength{\tabcolsep}{4.0pt}
\renewcommand{\arraystretch}{1.06}
\begin{adjustbox}{max width=\linewidth}
\begin{tabular}{lccccc}
\toprule
Reward variant & Eval. $N$ & Support $N$ & EM$\uparrow$ & Human support$\uparrow$ & Adv.--support corr.$\uparrow$ \\
\midrule
Score-permutation control & 600 & 180 & $0.350$ & $0.630\,[0.580,0.680]$ & $0.02\,[-0.13,0.17]$ \\
\rowcolor{gray!5}
Cost-only & 600 & 180 & $0.350$ & $0.660\,[0.610,0.710]$ & $0.08\,[-0.07,0.23]$ \\
\rowcolor{cyan!10}
Search-G1 & 600 & 180 & $0.370$ & $0.730\,[0.680,0.780]$ & $0.41\,[0.28,0.52]$ \\
\bottomrule
\end{tabular}
\end{adjustbox}
\end{table}

\end{document}